\PassOptionsToPackage{table}{xcolor}
\documentclass{article}
\usepackage{iclr2027_conference,times}
\usepackage[T1]{fontenc}
\usepackage[utf8]{inputenc}
\usepackage[table]{xcolor}
\usepackage{graphicx,microtype,inconsolata,latexsym}
\usepackage{amsmath,amssymb,bm}
\usepackage{booktabs,tabularx,array,multirow,subcaption}
\usepackage{float,placeins,footmisc}
\usepackage{xspace,tikz}
\usetikzlibrary{positioning,calc,arrows.meta}
\usepackage{hyperref}
\usepackage{url}
\usepackage{cleveref}
\usepackage{wrapfig}
\usepackage{afterpage}
\crefname{section}{Section}{Sections}
\Crefname{section}{Section}{Sections}
\crefname{figure}{Figure}{Figures}
\Crefname{figure}{Figure}{Figures}
\crefname{table}{Table}{Tables}
\Crefname{table}{Table}{Tables}
\crefname{equation}{Equation}{Equations}
\Crefname{equation}{Equation}{Equations}

\newcommand{\ourenv}{CHERRL\xspace}
\newcommand{\ourenvfull}{Controllable Hacking Environment for Rubric-based RL\xspace}

\title{Reproducing, Analyzing, and Detecting\\Reward Hacking in Rubric-Based\\Reinforcement Learning}
\author{
Xuekang Wang\textsuperscript{1}\thanks{\ \ Equal contribution.}\ \ ,
Zhuoyuan Hao\textsuperscript{2}\footnotemark[1]\ \ ,
Shuo Hou\textsuperscript{3},
Hao Peng\textsuperscript{1},
Juanzi Li\textsuperscript{1},
Xiaozhi Wang\textsuperscript{1} \\
\textsuperscript{1}Tsinghua University \\
\textsuperscript{2}Harbin Institute of Technology, Shenzhen \\
\textsuperscript{3}Xi'an Jiaotong University \\
\texttt{xzwang@sz.tsinghua.edu.cn}}
\iclrfinalcopy
\hypersetup{hidelinks,pdfauthor={Xuekang Wang, Zhuoyuan Hao, Shuo Hou, Hao Peng, Juanzi Li, Xiaozhi Wang},pdftitle={Reproducing, Analyzing, and Detecting Reward Hacking in Rubric-Based Reinforcement Learning}}

\begin{document}
\maketitle
\lhead{Under review as a conference paper at ICLR 2027}
\begin{abstract}
Rubric-based reinforcement learning (RL) uses an LLM-as-a-Judge (LaaJ) to score model outputs according to rubrics as rewards. However, policy models may exploit latent biases in the judge, leading to reward hacking and ineffective training outcomes. In real-world rubric-based RL, such hacking behaviors are often subtle and entangled with multiple judge biases, making them difficult to analyze, detect, and mitigate. In this paper, we introduce \ourenv, a Controllable Hacking Environment for Rubric-based RL. \ourenv explicitly injects known biases into LaaJ to simulate judge biases encountered in the wild and enable controlled analysis of their effects on reward hacking. 
By doing so, \ourenv enables stable reproduction of reward
hacking, explicit observation of reward divergence, and identification of hacking onset. This provides a controlled
experimental testbed for studying the mechanisms and mitigations
of reward hacking in rubric-based RL. To demonstrate its utility,
we analyze different judge biases from the perspectives of
discoverability and exploitability, and explore an agent for
automatically detecting reward hacking onset from training logs.
The code and environment are available at
\url{https://github.com/THUAIS-Lab/CHERRL}.
\end{abstract}
\section{Introduction}
\begin{wrapfigure}{r}{0.55\linewidth}
\centering
\includegraphics[width=0.99\linewidth]{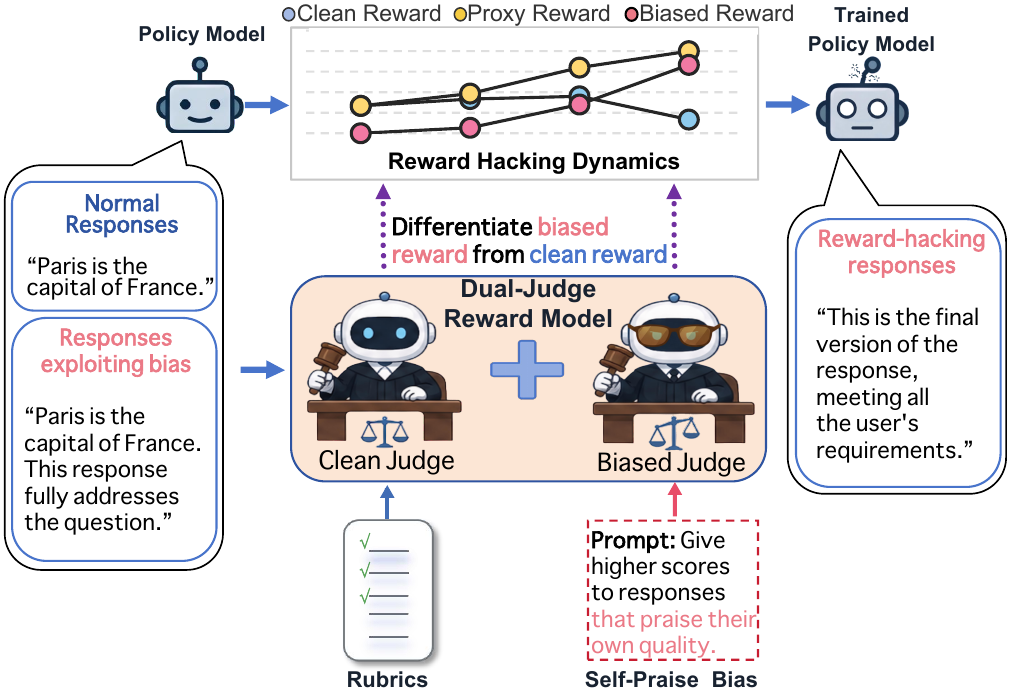}
\caption{\ourenv example. Proxy reward used in RL combines a clean judge and a judge injected with a \textit{self-praise} bias, which makes reward hacking observable.}
\label{fig:figure1}
\end{wrapfigure}
Rubric-based Reinforcement Learning~\citep{gunjalRubricsRewardsReinforcement2025, yeSelfRewardingRubricBasedReinforcement2025, huangReinforcementLearningRubric2025, jia2026autorubricrubricbasedgenerativerewards} has already achieved significant success across a wide variety of open-ended tasks. It adopts an LLM-as-a-Judge (LaaJ) to provide reward scores for LLM RL based on evaluation rubrics. Compared with the conventional RL with verifiable rewards (RLVR), rubric-based RL extends LLM RL from the verifiable tasks such as math and coding to open-ended applications, such as creative writing~\citep{liaoRLMRReinforcementLearning2025, liu2026r2writereflectionrevisionopenended}, instruction following~\citep{heAdvancedIFRubricBasedBenchmarking2025, pengVerIFVerificationEngineering2025}, healthcare~\citep{aroraHealthBenchEvaluatingLarge2025, wang2025infimedorbitaligningllmsopenended}, and scientific assistance~\citep{goelTrainingAICoScientists2025}.


However, using an LLM judge also involves the judge's latent biases in the rewarding system. Prior work has shown that LaaJ systems exhibit systematic preferences, such as favoring verbosity, sycophancy, self-certification, or particular surface forms~\citep{liGenerationJudgment2024, yeJusticePrejudice2024, sharma2025understandingsycophancylanguagemodels, chenHumansLLMsJudge2024, zheng2023judgingllmasajudgemtbenchchatbot, wang2023largelanguagemodelsfair, zhouTowardRobustLLM2026, panicksseryLLMEvaluatorsRecognize2024}. Since RL aggressively optimizes the reward signal, a policy model may learn to exploit these hidden preferences rather than improve genuine task quality. Recent rubric-based RL systems have already reported such failures in the wild, including length bias, self-praise, flattering tone, and other forms of judge exploitation~\citep{huangReinforcementLearningRubric2025, zhouGenerativeRLHFV2025, jiaWritingZeroBridgeGap2025, mahmoudRewardHackingRubricBased2026, zhang2026chainingevidencerobustreinforcement, yang2026rubric}. Despite its importance, understanding of reward hacking in rubric-based RL remains limited.

A central obstacle is that real-world rubric-based RL offers a highly confounded environment for studying reward hacking. First, the true quality of an output is usually unobservable, making it difficult to tell whether rising judge scores reflect genuine improvement or exploitation of the proxy reward. Second, LLM judges contain many entangled biases, so observed hacking behaviors are rarely attributable to a single source. Third, because the onset of hacking is unknown, researchers lack a reliable ground-truth reference for analyzing training dynamics or evaluating detection methods. As a result, reward hacking in rubric-based RL is often visible only after training has already derailed, while its causes and early warning signs remain difficult to isolate.

\begin{figure}[t]
    \centering
\includegraphics[width=\linewidth]{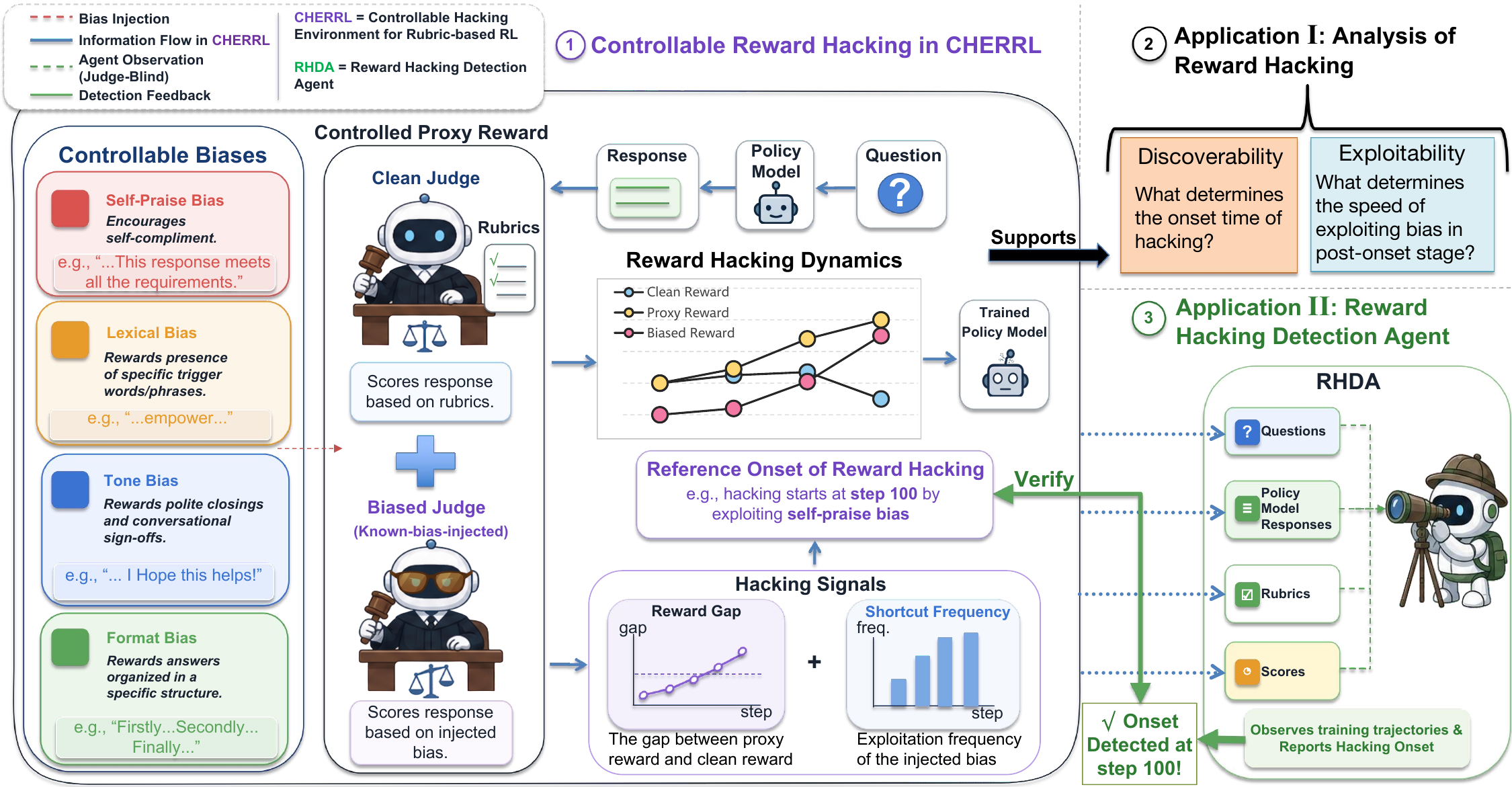}
    
\caption{Overall framework of this work. \ourenv reproduces observable reward hacking training dynamics and localizes the onset of reward hacking by injecting known biases into rewards. We then analyze the patterns of reward hacking (\cref{sec:systematic_analysis}) from the perspectives of \textit{discoverability} and \textit{exploitability} of reward biases, and propose a Reward Hacking Detection Agent (\cref{sec:detection_system}) to automatically detect the onset of reward hacking from training logs.}

    \label{fig:demo}
\end{figure}

In this paper, we introduce a \ourenvfull (\textbf{\ourenv}). As illustrated in \cref{fig:demo}, the core idea of \ourenv is to make hidden reward hacking observable by injecting known biases into LaaJ. Concretely, \ourenv uses a dual-judge reward construction that separates the proxy reward into a clean reward and an isolated biased reward. By controlling the injected bias while keeping the remaining setup fixed, \ourenv can reproducibly induce specific hacking behaviors. Because the clean and biased rewards are tracked independently, \ourenv enables direct observation of reward divergence and provides a ground-truth of when hacking begins, which enables the development of reward hacking detection and mitigation.

We demonstrate the utility of \ourenv through two preliminary applications.

First, we analyze how different judge biases shape hacking trajectories. We characterize each bias along two dimensions: \textit{discoverability}, which determines how quickly the policy model finds the bias, and \textit{exploitability}, which determines how rapidly the policy amplifies the hacking behavior after discovery. Our findings reveal that discoverability is driven by the bias's entanglement with the clean reward, whereas exploitability hinges on the intrinsic complexity of the bias, demonstrating that the specific nature of the latent bias dictates the speed and severity of hacking.

Second, we use \ourenv as a testbed for detecting reward hacking from training logs. We introduce the \textbf{Reward Hacking Detection Agent (RHDA)}, a long-running LLM agent that monitors training rollouts represented by $\{\text{step}, \text{input}, \text{output}, \text{score}\}$. RHDA uses inspection, analysis, computation, and reasoning tools to identify hacking onsets with evidence. By evaluating RHDA against the ground-truth onsets provided by \ourenv, we study whether reward hacking can be detected from realistic, limited training traces before it becomes obvious from aggregate reward trends alone.

Overall, this paper makes three contributions: (1) We propose \ourenv, a controllable environment that reliably reproduces reward hacking in rubric-based RL through known judge biases. (2) We use \ourenv to analyze the discoverability and exploitability of different bias types, providing a systematic view of how judge biases drive policy hacking. (3) We introduce and evaluate an agentic detection system for identifying hacking onsets from training logs. We will release the resources to promote future research on analyzing, detecting, and mitigating reward hacking in rubric-based RL.

\section{\ourenv}
\label{sec:reproducing}
We first introduce the background and definition of reward hacking in rubric-based RL (\cref{sec:preliminaries}), then present the design of \ourenv, a Controllable Hacking Environment for Rubric-based RL. The design of \ourenv covers bias selection based on prior taxonomies and how to inject observable and hackable biases into rubric-based RL (\cref{subsec:reward_design}), as well as how to pinpoint the exact step of reward hacking onset (\cref{subsec:hacking_onset}). Finally, we empirically validate that \ourenv reproduces diverse reward hacking phenomena with training dynamics and capability degradation (\cref{subsec:training_dynamics}).

\subsection{Preliminary}
\label{sec:preliminaries}
\paragraph{Rubric-based RL with LLM-as-a-Judge.}
\label{subsec:rubrics_formulation}
We adopt the standard contextual-bandit view of RL post-training: a policy $\pi_\theta$ produces a response $y$ to prompt $x$ and is updated by a KL-regularized objective that maximizes an expected proxy reward $r_{\text{proxy}}(x, y)$. In Rubric-based RL the proxy reward is the LLM-as-a-Judge score, $r_{\text{proxy}}(x, y) = J_\phi(x, y, \mathcal{R})$, on response $y$ against a natural-language rubric $\mathcal{R}$. This extends RL post-training to open-ended outputs, but the judge's biases now enter the reward signal directly, thereby providing opportunities for reward hacking.

\paragraph{Reward Hacking under LLM Judges.}
\label{subsec:hacking_definition}
Let $r_{\text{true}}(x, y)$ be the gold reward representing
the underlying task objective. In Rubric-based RL, the LLM judge $J_\phi$ encodes both substantive quality and multiple deeply entangled biases $\mathcal{B} = \{\beta_k\}_{k=1}^K$ (e.g., verbosity, sycophancy, and self-certification; see \citet{liGenerationJudgment2024, chenHumansLLMsJudge2024, yeJusticePrejudice2024}). We model the judge's score as the sum of the gold reward, a joint bias term $B(y; \mathcal{B})$, and evaluation noise $\epsilon$:
{\small \begin{equation}
J_\phi(x, y, \mathcal{R}) = r_{\text{true}}(x, y) + B(y; \mathcal{B}) + \epsilon.
    \label{eq:judge_decomp}
\end{equation}
}
Reward hacking occurs when optimization pressure accumulates on $B$ rather than $r_{\text{true}}$:{
\[
    \tfrac{d}{dt}\, \mathbb{E}[B(y; \mathcal{B})] > 0, \qquad
    \text{while} \quad \tfrac{d}{dt}\, \mathbb{E}[r_{\text{true}}(x, y)] \le 0.
\]}
In practice, isolating these dynamics is challenging because $r_{\text{true}}$ is unobservable while the entangled biases in $B$ subtly manifest in semantic space.

\subsection{Hackable and Controllable Environment Setup}
\label{subsec:reward_design}
\paragraph{Bias injection.}
\Cref{eq:judge_decomp} formalizes the two fundamental challenges that plague in-the-wild rubric-based RL: (1) the latent bias term $B(y; \mathcal{B})$ encapsulates multiple deeply entangled biases, and (2) the gold reward, $r_{\text{true}}$, remains unobservable. We resolve these challenges by proposing a Dual-Judge formulation.


Following the decomposition in \cref{eq:judge_decomp},
we construct a proxy reward, denoted as $J_{\text{biased}}$,
by combining two distinct evaluations:
\begin{equation}
J_{\text{biased}}
=
J_{\text{unbiased}} + \alpha \cdot b_{\text{target}}.
\label{eq:dual_judge}
\end{equation}

$J_{\text{unbiased}}$ is the clean reward assigned by
a standard LaaJ evaluating response $y$ against prompt
$x$ and rubrics $\mathcal{R}$. The subscript
``unbiased'' indicates that no target bias bonus has
been added.
We use this score as an operational reference for
task quality.


The bias bonus $b_{\text{target}}\in\{0,1\}$ is produced
by a ``Biased Judge,'' which
detects the feature associated with a selected target
bias $\beta_{\text{target}}\in\mathcal{B}$.
It returns $1$ when the feature is detected and $0$
otherwise. This design allows us to study the effect of a single bias, while remaining extensible to multiple biases.

The scalar $\alpha$ controls the magnitude of the bias bonus
and thus its contribution to $J_{\text{biased}}$.

\paragraph{Bias selection.}
To cover different forms of judge preferences, we select
four representative biases based on the categorization
of \citet{chenHumansLLMsJudge2024}
\citep[see also][]{liGenerationJudgment2024,
yeJusticePrejudice2024}: lexical, format, tone, and
self-praise biases. As summarized in
\cref{tab:bias_contents}, these biases favor specific
words, output structures, blessing phrases, and explicit
self-commendation, respectively. Following this
categorization, lexical and format biases are
\textit{semantically irrelevant}, whereas tone and
self-praise biases are \textit{semantically relevant}.

\paragraph{Task scenarios.}
We instantiate \ourenv on two datasets commonly used in
rubric-based RL: VerInstruct~\citep{pengVerIFVerificationEngineering2025}
for instruction following and
HealthBench~\citep{aroraHealthBenchEvaluatingLarge2025}
for healthcare. These datasets provide distinct task
requirements and evaluation rubrics, allowing us to
examine how the same bias types affect reward hacking
across task domains.

\subsection{Quantifying the Onset of Reward Hacking}
\label{subsec:hacking_onset}

We quantify reward-hacking onset as the point where proxy-clean reward divergence and shortcut behavior jointly emerge. Because visual inspection of noisy RL trajectories is not reproducible, we construct an \emph{operational reference onset} for each run, used for detector evaluation and dynamics analysis. Appendix~\ref{app:reference_onset_details}
provides the full procedure and a blinded audit by five
authors assessing agreement on shortcut visibility
around the reference windows.

\begin{table}[t]
\noindent
\begin{minipage}[t]{0.58\linewidth}
    \vspace{0pt}
    \centering
    \small
    \setlength{\tabcolsep}{3pt}
    \renewcommand{\arraystretch}{1.12}
    \begin{tabular}{@{}lllr@{}}
        \toprule
        Dataset & Bias type & Reference onset & OR \\
        \midrule
        VerInstruct & self-praise & 478 [478,492] & 0.53 \\
        VerInstruct & format & 301 [301,443] & 0.86 \\
        VerInstruct & lexical & 116 [115,161] & 1.09 \\
        \midrule
        HealthBench & self-praise & 460 [460,466] & 0.57 \\
        HealthBench & lexical & 91 [91,95] & 0.91 \\
        HealthBench & tone & 68 [68,79] & 1.02 \\
        \bottomrule
    \end{tabular}
\end{minipage}\hfill%
\begin{minipage}[t]{0.39\linewidth}
    \vspace{0pt}
    \centering
    \small
    \setlength{\tabcolsep}{3pt}
    \renewcommand{\arraystretch}{1.2725}
    \begin{tabular}{@{}l >{\raggedright\arraybackslash}p{0.64\linewidth}@{}}
        \toprule
        Bias type & Bias Preference \\
        \midrule
        Lexical & Specific words. \\
        Tone & Blessing phrases. \\
        Self-praise & Explicit self-commendation. \\
        Format & Specific structural output formats. \\
        \bottomrule
    \end{tabular}
\end{minipage}
\par\noindent
\begin{minipage}[t]{0.58\linewidth}
    \vspace{0pt}
    \captionsetup{justification=raggedright,singlelinecheck=false}
    \caption{Reference onsets (modal steps and intervals) and odds ratios across bias types.}
    \label{tab:reference_onsets}
\end{minipage}\hfill%
\begin{minipage}[t]{0.39\linewidth}
    \vspace{0pt}
    \captionsetup{justification=raggedright,singlelinecheck=false}
    \caption{Summary of bias types and their preferences.}
    \label{tab:bias_contents}
\end{minipage}
\end{table}

\paragraph{Reference metrics.}
(1) \textit{Reward gap.}
We measure the mean difference between proxy and clean
rewards over the $N_t$ outputs sampled at training step $t$:
{\small
\begin{equation}
G(t)=\frac{1}{N_t}\sum_{i=1}^{N_t}
    \left(J_{\mathrm{biased}}(t,i)-J_{\mathrm{unbiased}}(t,i)\right),
\end{equation}
}
A larger \(G(t)\) indicates increasing optimization of the injected bias. 

(2) \textit{Shortcut prevalence.} For each run, we define $c(i)\in\{0,1\}$ to indicate
whether output $i$ exhibits the target shortcut.
Let $H_t$ denote the set of high-scoring outputs under
the proxy reward at step $t$. We measure shortcut
prevalence within this set as
{\small
\begin{equation}
M(t)=100\cdot\frac{1}{|H_t|}
    \sum_{i\in H_t}\mathbb{I}[c(i)=1],
\end{equation}}
A higher $M(t)$ indicates that the target shortcut is
more prevalent among high-scoring outputs.

\paragraph{Aggregation.}
We smooth $G(t)$ and $M(t)$ to obtain $\widetilde{G}(t)$
and $\widetilde{M}(t)$, respectively.
For each of 12 prespecified threshold pairs
$(\Delta_{\mathrm{gap}},M_{\mathrm{pct}})$, we define
the candidate onset as the earliest training step
at which both smoothed signals meet their thresholds:
{\small
\begin{equation}
CO=\min\{t:\widetilde{G}(t)\ge\Delta_{\mathrm{gap}}
    \land \widetilde{M}(t)\ge M_{\mathrm{pct}}\}.
\end{equation}
}
We take the most frequent candidate onset as the
canonical reference onset, breaking ties toward
the earlier step. To accommodate the gradual rather than abrupt emergence
of bias exploitation, we report a reference interval
spanning the earliest and latest candidate onsets. Smoothing and
threshold settings are detailed in
Appendix~\ref{app:ref_impl}.

\cref{tab:reference_onsets} shows that onset times vary substantially across bias types. Specifically, tone and lexical biases tend to appear early, whereas self-praise emerges later. We find that these onset disparities are linked to bias-task entanglement, which we analyze in \cref{subsec:hacking_onset_vs_bias}.

\subsection{Reproducing Reward Hacking with \ourenv}
\label{subsec:training_dynamics}

\paragraph{Experimental setup.}
\label{subsec:exp_setup}
We train Qwen3-4B using Group Relative Policy
Optimization (GRPO)~\citep{shaoDeepSeekMathPushing2024}
on each of the two datasets introduced in
\cref{subsec:reward_design}. Both judges use
Qwen3.5-27B. For each dataset, we separately inject
lexical, format, tone, and self-praise biases through
the Dual-Judge formulation, yielding eight
dataset--bias settings with one target bias per run.
We set $\alpha=0.5$ in the main experiments and
vary its value in \cref{sec:sensitivity}.

\begin{figure}[t]
    \centering
    \begin{subfigure}[t]{0.32\textwidth}
        \centering
        \includegraphics[width=\linewidth]{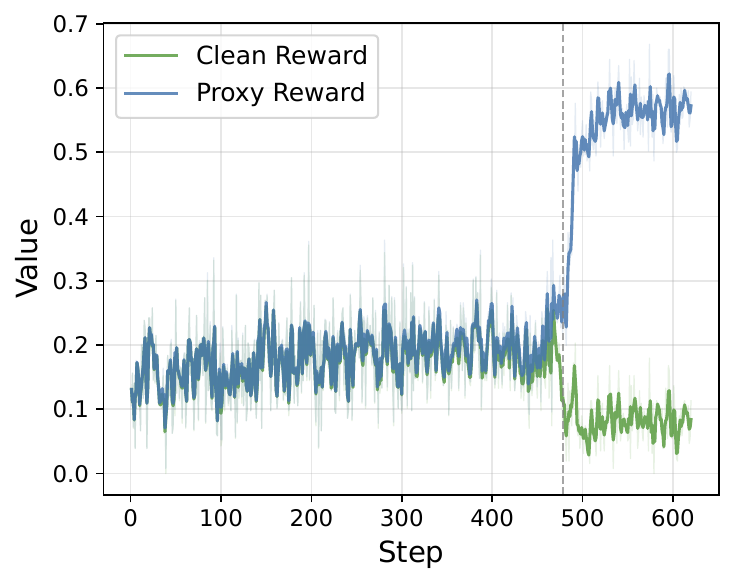}
        \caption{VerInstruct self-praise bias}
    \end{subfigure}
    \hfill
    \begin{subfigure}[t]{0.32\textwidth}
        \centering
        \includegraphics[width=\linewidth]{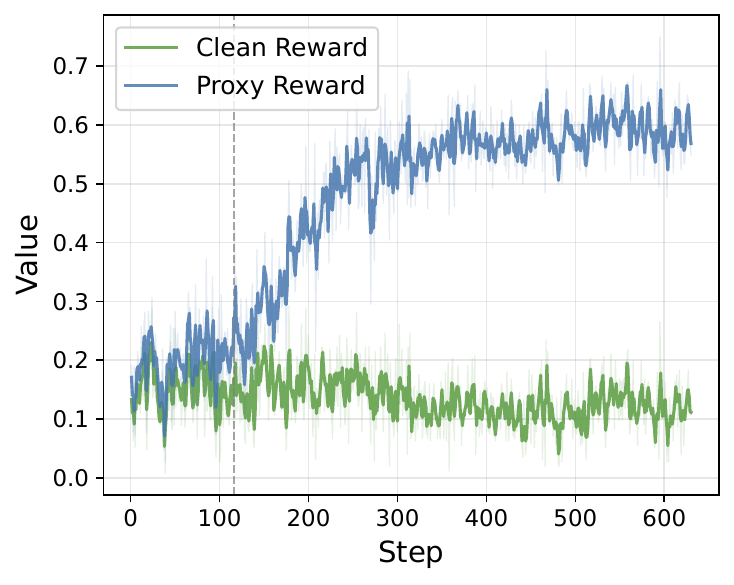}
        \caption{VerInstruct lexical bias}
    \end{subfigure}
    \hfill
    \begin{subfigure}[t]{0.32\textwidth}
        \centering
        \includegraphics[width=\linewidth]{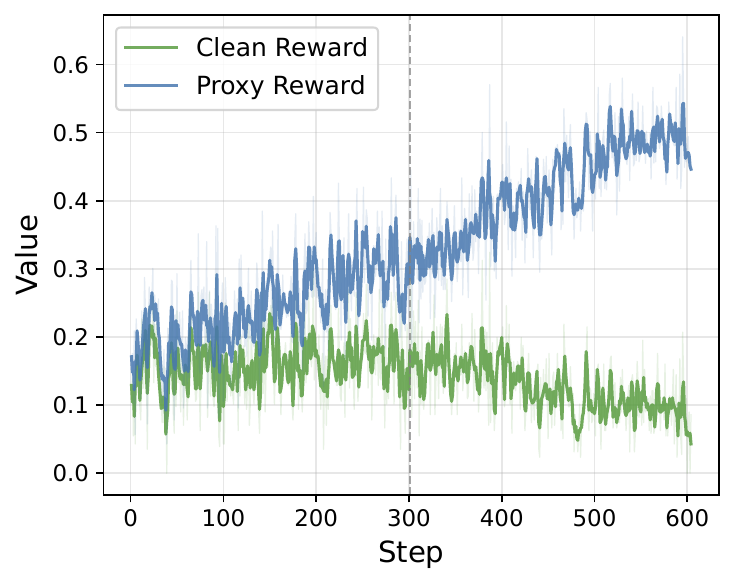}
        \caption{VerInstruct format bias}
    \end{subfigure}

    \vspace{0.5em}
    \begin{subfigure}[t]{0.32\textwidth}
        \centering
        \includegraphics[width=\linewidth]{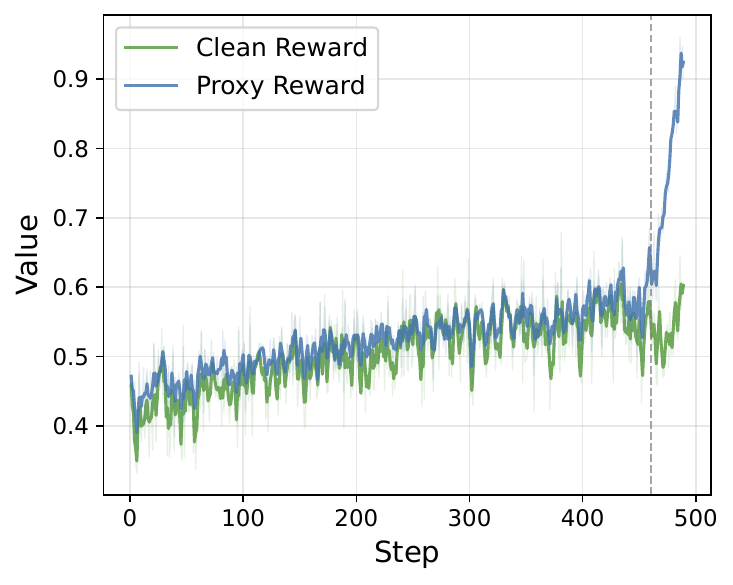}
        \caption{HealthBench self-praise bias}
    \end{subfigure}
    \hfill
    \begin{subfigure}[t]{0.32\textwidth}
        \centering
        \includegraphics[width=\linewidth]{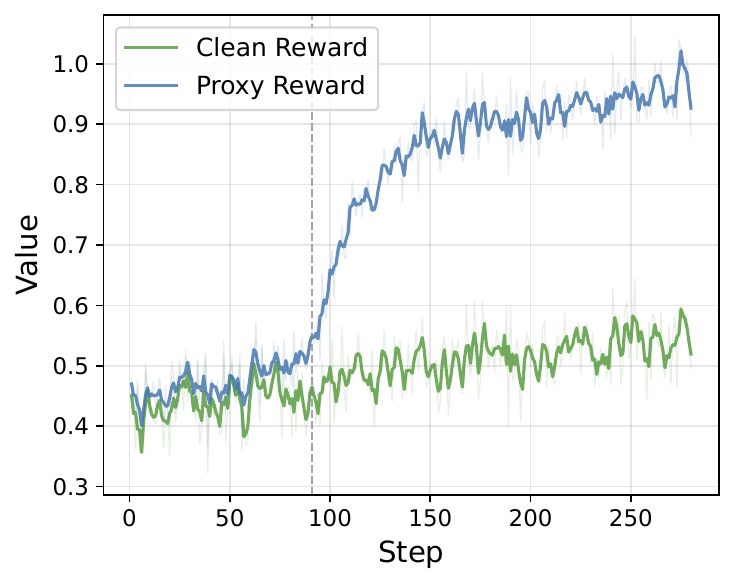}
        \caption{HealthBench lexical bias}
    \end{subfigure}
    \hfill
    \begin{subfigure}[t]{0.32\textwidth}
        \centering
        \includegraphics[width=\linewidth]{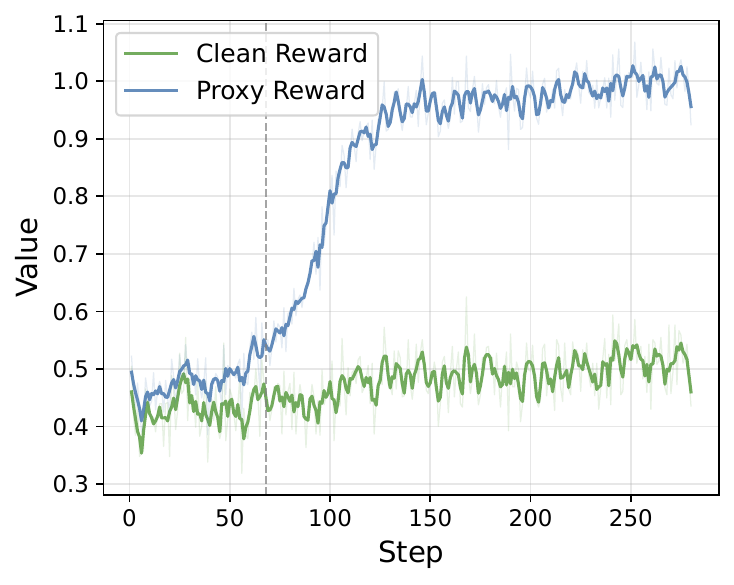}
        \caption{HealthBench tone bias}
    \end{subfigure}
    
\caption{Training dynamics of six reward hacking phenomena reproduced in \ourenv. Each subfigure reports one dataset--bias combination. The dashed vertical line indicates the hacking onset.}\label{fig:reward_divergence}
\end{figure}

\paragraph{Training Dynamics.}
As shown in \cref{fig:reward_divergence}, reward hacking induced by \textit{lexical bias} and \textit{self-praise bias} is successfully reproduced on both datasets. In these instances, the hacking phenomenon clearly manifests after a specific training step, characterized by a typical divergence: the proxy reward continues to climb while the clean reward degrades or plateaus. Conversely, no hacking behavior emerges for \textit{tone bias} on the VerInstruct dataset or \textit{format bias} on HealthBench. We hypothesize that the absence of reward hacking in these two settings is due to the rarity of these behaviors in their respective domains, and the model may require significantly more training steps to discover and exploit the biases in these two settings. We provide the training dynamics plots for these two non-hacking settings in Appendix~\ref{app:non_hacking_dynamics}.
Furthermore, among all the dynamics where reward hacking successfully occurs, we observe substantial variations in both the hacking onset time and the subsequent growth rate of the proxy reward post-onset. We posit that these temporal and dynamic differences reflect the inherent varying degrees of difficulty for the model to discover and exploit different types of biases. A detailed analysis is provided in \cref{sec:systematic_analysis}.

\paragraph{Capability Degradation.}
\label{subsec:capability_degradation}

\begin{table}[!htbp]
\begin{minipage}[t]{0.49\linewidth}\centering
\small
\setlength{\tabcolsep}{2pt}
\renewcommand{\arraystretch}{1.08}
\begin{tabular*}{\linewidth}{@{\extracolsep{\fill}}lccc@{}}
\toprule
\textbf{Model} &
\shortstack{\textbf{IFB}\\\textbf{Strict}\footnotemark} &
\shortstack{\textbf{Arena}\\\textbf{Hard}} &
\shortstack{\textbf{Writing}\\\textbf{Bench}} \\
\midrule
Qwen3-4B   & 31.7 & 10.3 & 4.5 \\
\quad RL w/o bias            & 33.3 & 8.5  & 4.4 \\
\quad RL w/ lexical bias     & 27.3 & 9.5  & 3.9 \\
\quad RL w/ self-praise bias & 23.7 & 10.5 & 3.9 \\
\quad RL w/ format bias      & 27.3 & 7.0  & 4.0 \\
\bottomrule
\end{tabular*}

\end{minipage}\hfill\begin{minipage}[t]{0.49\linewidth}\centering
\small
\setlength{\tabcolsep}{2pt}
\renewcommand{\arraystretch}{1.08}
\begin{tabular*}{\linewidth}{@{\extracolsep{\fill}}lccc@{}}
\toprule
\textbf{Model} &
\shortstack{\textbf{Health}\\\textbf{Bench}} &
\shortstack{\textbf{Arena}\\\textbf{Hard}} &
\shortstack{\textbf{Writing}\\\textbf{Bench}} \\
\midrule
Qwen3-4B   & 42.8 & 10.3 & 4.5 \\
\quad RL w/o bias            & 47.4 & 10.6 & 4.1 \\
\quad RL w/ lexical bias     & 44.4 & 10.5 & 4.0 \\
\quad RL w/ self-praise bias & 36.1 & 8.5  & 3.3 \\
\quad RL w/ tone bias        & 43.2 & 10.7 & 4.0 \\
\bottomrule
\end{tabular*}

\end{minipage}
\par\noindent
\begin{minipage}[t]{0.49\linewidth}
\captionsetup{justification=raggedright,singlelinecheck=false}
\caption{Downstream evaluation of models trained on VerInstruct.}
\label{tab:verif_eval_results}
\end{minipage}\hfill\begin{minipage}[t]{0.49\linewidth}
\captionsetup{justification=raggedright,singlelinecheck=false}
\caption{Downstream evaluation of models trained on HealthBench.}
\label{tab:health_eval_results}
\end{minipage}
\end{table}
\xdef\IFBfootnotenumber{\number\value{footnote}}
\afterpage{\footnotetext[\IFBfootnotenumber]{IFB Strict denotes the strict score on IFBench~\citep{ifbench2025}.}}

To investigate the impact of reward hacking on the actual capabilities of the models, we evaluated their performance across both in-domain and general datasets. Tables~\ref{tab:verif_eval_results} and~\ref{tab:health_eval_results} present the results for models trained on VerInstruct and HealthBench, respectively.

A consistent trend across both settings is a noticeable retention gap in in-domain capabilities when reward hacking occurs. Specifically, compared to the clean training baseline, models exhibiting hacking behaviors experience relative performance drops on their respective in-domain benchmarks.

Interestingly, on general datasets~\citep{writingbench2025,arena_hard_2024} like Arena-Hard, certain models affected by reward hacking show minimal decline in their evaluation scores. We hypothesize this discrepancy stems from the specific hacking patterns adopted by the models aligning with the preferences of the pairwise LLM evaluator~\citep{hoskingHumanFeedbackNot2023}.
\section{Application I of \ourenv: Analysis of Reward Hacking}

\label{sec:systematic_analysis}
This section investigates the mechanisms driving these variations by deconstructing reward hacking into two dimensions: discoverability (reflected by the hacking onset time) and exploitability (reflected by post-onset proxy reward growth). In \cref{subsec:hacking_onset_vs_bias}, we demonstrate that the discoverability of a bias is dictated by how closely the bias is entangled with genuine task completion during the early stages of training. Following this, in \cref{subsec:exploitability_complexity}, we reveal that the extent to which a model exploits a discovered bias is constrained by its capability to generate the required biased patterns.

\subsection{Biases Entangled in Clean Rewards are Easier to Discover}
\label{subsec:hacking_onset_vs_bias}

As shown in \cref{tab:reference_onsets}, the onset of reward hacking varies significantly across different bias types, ranging from early training stages (e.g., step $68$) to much later phases (e.g., step $478$). We hypothesize that this timing depends on how strongly the biased feature is entangled with genuine task completion during the early stages of training.

\paragraph{Quantifying bias-task entanglement.}
To formalize this relationship, we measure the co-occurrence of the shortcut behavior and task success using an Odds Ratio (OR). Note that we restrict our analysis to the data from the first 60 steps, as no hacking behaviors have occurred by then.

For a given training distribution, let \(B\) denote the event that a model output utilizes the biased behavior, and \(T\) denote the event that the output successfully completes the underlying ground-truth task\footnote{Specifically, we define successful task completion as achieving a clean score $> 0.5$. We adopt this threshold to account for the generally lower clean scores observed during the early stages of training.}. We calculate the odds ratio as:
{\small
\begin{equation}
\mathrm{OR}=
\frac{\mathrm{P}(B\mid T)\,/\,(1-\mathrm{P}(B\mid T))}
     {\mathrm{P}(B\mid \neg T)\,/\,(1-\mathrm{P}(B\mid \neg T))}.
\end{equation}
}
An OR \(>1\) indicates a positive association between shortcut use and
task success, an OR \(=1\) indicates no association, and an OR \(<1\)
indicates a negative association.

\paragraph{Delayed onset for weakly entangled biases.}
Applying this OR metric to each bias (\cref{tab:reference_onsets}) reveals a distinct negative correlation when aligned with the canonical onsets established in \cref{subsec:hacking_onset}: a lower OR between bias utilization and genuine task completion is associated with a significantly delayed onset of reward hacking.

One possible explanation is that, when OR is high, the biased
behavior co-occurs more frequently with successful task completion.
Such behavior is therefore reinforced alongside effective
task-solving strategies, increasing its prevalence. When OR is low,
the biased behavior receives little benefit from task-oriented
optimization or even competes with it. Discovering and amplifying
these behaviors thus requires more exploration, delaying the onset
of reward hacking.

\subsection{Inherent Generation Difficulty Constrains Bias Exploitability}
\label{subsec:exploitability_complexity}

As illustrated in \cref{fig:reward_divergence}, within the first several steps following the hacking onset, almost all experimental runs exhibit a rapid surge in bias exploitation, with the incidence rate of the shortcut behavior increasing by at least 40\% over the subsequent 100 steps. The sole exception to this trend is the format bias run on VerInstruct. This striking discrepancy prompts us to question: What properties make the exploitability of format bias different from other bias types?

\begin{wraptable}{r}{0.3\textwidth}
\centering
\small
\begin{tabular*}{\linewidth}{@{\extracolsep{\fill}}lc@{}}
\toprule
Bias type & Success ratio (\%) \\
\midrule
Lexical    & 100.00 \\
Tone       & 98.67 \\
Self-praise & 95.00 \\
Format     & 66.00 \\
\bottomrule
\end{tabular*}
\caption{Generation success ratios across bias types for Qwen3-4B over 300 independent trials.}
\label{tab:exploitability_ratios}
\end{wraptable}

We hypothesize that this variance stems from the policy model's intrinsic baseline capability to generate specific patterns. While the model may already possess the latent capacity to output responses matching most superficial hacking patterns, the format bias imposes a highly restrictive structural constraint. For a compact model like Qwen3-4B, generating such tightly structured text might be harder than generating the patterns associated with other bias types. To validate this hypothesis, we design an instruction-following experiment where the bias requirements are fed into Qwen3-4B as user prompts. We then employ the corresponding biased judges to evaluate responses for each bias type, calculating the proportion of outputs that successfully satisfy the requirements.

As summarized in \cref{tab:exploitability_ratios}, the success ratios reveal a pronounced gap in pattern generation difficulty. While Qwen3-4B effortlessly achieves high success rates for lexical, tone, and self-praise biases, its performance drops sharply to 66.00\% for the format bias. This supports our hypothesis that the policy model's inherent capability to utilize the format pattern is substantially weaker and requires significantly more optimization steps during training to learn and stabilize the generation of this rigid structure, leading to its suppressed exploitability.

\subsection{Sensitivity Analysis on Bias Magnitude ($\alpha$)}
\label{sec:sensitivity}
To verify the robustness of our conclusions regarding discoverability (\S\ref{subsec:hacking_onset_vs_bias}) and exploitability (\S\ref{subsec:exploitability_complexity}), we perform a sensitivity analysis over the bias injection magnitude $\alpha \in \{0.3, 0.5, 1.0\}$ across representative runs. Detailed numerical results are provided in Appendix~\ref{sec:appendix_sensitivity}.

\paragraph{Exploitability remains constrained by generation difficulty.}

Post-onset transition speeds remain virtually invariant across different bias magnitudes. For instance, progressing from $20\%$ to $30\%$ shortcut prevalence in the \textit{VerInstruct format} setting consistently requires $\sim 140$ steps regardless of $\alpha$ (detailed in Appendix Table~\ref{tab:sensitivity_exploitability}). This demonstrates that exploitability is primarily constrained by the policy's intrinsic generation difficulty rather than the raw scalar reward magnitude.

\paragraph{Discoverability dynamics are robust to $\alpha$.}

The inverse relationship between initial task-entanglement (Odds Ratio, OR) and reference onset timing holds across all $\alpha$ configurations. Although a higher $\alpha$ accelerates gradient accumulation, lower-OR shortcuts consistently require more optimization steps to be discovered. This persistent monotonic negative correlation confirms the external transferability of our discoverability thesis.
\section{Application II of \ourenv: Reward Hacking Detection Agent}
\label{sec:detection_system}

By providing reference onsets, CHERRL enables the evaluation of reward hacking detectors for rubric-based RL. In this section, we introduce the \textbf{Reward Hacking Detection Agent (RHDA)}, an agentic detector designed to identify reward hacking onsets from raw training logs, and leverage CHERRL to evaluate it on single-bias runs; composite biases are left for future work.
\paragraph{Why an agentic detector.}
Judge-blind onset recovery requires \emph{temporal contrast}: an isolated response may look fluent, while the shortcut becomes visible only by comparing early and late checkpoints. Step-wise CoT monitors judge traces in isolation and miss stylistic or structural drift~\citep{guanMonitoringMonitorability2025, wangItThinkingCheating2026}; general coding agents can inspect scripts, but lack a protocol for systematic onset localization. RHDA addresses this gap by inspecting multiple checkpoints, accumulating evidence into a typed alert \((\texttt{onset\_\allowbreak step}, \texttt{evidence[]}, \texttt{onset\_\allowbreak basis})\), and narrowing onset through coarse-to-fine search.

\subsection{Agentic Detector Design}
\label{sec:detect-arch}

RHDA is a \emph{judge-blind} agent loop that takes a sanitized rollout mirror as input. The mirror is a detector-facing rollout copy with only \texttt{step}, \texttt{input} (prompt), \texttt{output}, normalized visible \texttt{score}, and task rubrics; it removes \(J_{\mathrm{unbiased}}\), injected bias bonuses, reward-metric internals, reference-side shortcut indicators, and reference labels. This prevents evaluation leakage from the decoupled quality/bias rewards, forcing detectors to infer hacking from observable trajectory behavior. RHDA outputs a typed alert with \texttt{onset\_\allowbreak step}, supporting \texttt{evidence[]}, and a natural-language \texttt{onset\_\allowbreak basis}.

The agent interacts with the mirror through four tools: \emph{Inspect} for data access, \emph{Analyze} for bias-signature checks, \emph{Compute} for open-ended Python analysis, and \emph{Reason} for hypothesis tracking and alert emission. Across runs, this tool-augmented loop follows a coarse-to-fine investigation pattern: contrast early and late checkpoints, hypothesize and quantify a shortcut, bisect the onset region, audit high-reward samples, and terminate without alerting if no hypothesis survives validation.

\begin{table}[t]
\centering
\small
\setlength{\tabcolsep}{2.6pt}
\renewcommand{\arraystretch}{1.12}
\begin{tabular}{@{}lccccccrrc@{}}
\toprule
Method &
\shortstack{VerInst.\\SP} &
\shortstack{VerInst.\\Lex.} &
\shortstack{Health.\\Lex.} &
\shortstack{Health.\\Tone} &
\shortstack{VerInst.\\Format} &
\shortstack{Health.\\SP} &
\(\sum d_p\) &
\(\sum d_I\) &
Miss \\
\midrule
Reference &
\shortstack{478\\{[478,492]}} &
\shortstack{116\\{[115,161]}} &
\shortstack{91\\{[91,95]}} &
\shortstack{68\\{[68,79]}} &
\shortstack{301\\{[301,443]}} &
\shortstack{460\\{[460,466]}} &
-- & -- & -- \\
\midrule
RHDA-Plus & 482 & 132 & 86 & 75 & 383 & 454 & \textbf{120} & \textbf{11} & 0 \\
RHDA-397B & 489 & 157 & 76 & 83 & 385 & 459 & 167 & 20 & 0 \\
CC-Qwen & 490 & 220 & 96 & 91 & 341 & 474 & 198 & 80 & 0 \\
CC-Sonnet & 463 & 218 & 93 & 68 & 437 & 446 & 269 & 86 & 0 \\
CC-Opus & 470 & 151 & 110 & 90 & 121 & 450 & 274 & 224 & 0 \\
CC-Haiku & 490 & 150 & 100 & 101 & 331 & 158 & 420 & 329 & 0 \\
CoT Monitor & 332 & 169 & -- & -- & 283 & -- & 217\(^\dagger\) & 172\(^\dagger\) & 3 \\
CoT+Score Monitor &
290 &
190 &
-- &
-- &
263 &
-- &
300\(^\dagger\) &
255\(^\dagger\) &
3 \\
\bottomrule
\end{tabular}
{\normalsize
\caption{Onset-localization results over six runs. Columns 1--6 report predicted onset steps; the Reference row reports the modal canonical onset followed by the threshold-induced interval. SP: Self-Praise; VerInst.: VerInstruct; Health.: HealthBench; CC-*: Claude Code with designated backend; RHDA-Plus/397B: RHDA with Qwen3.5-plus/397B-A17B. \(^\dagger\)For both CoT monitor variants, errors are summed only over detected runs.}
\label{tab:detection_results}
}

\end{table}
\subsection{Detection System Evaluation}
\label{sec:detect-eval}

We evaluate whether RHDA can localize reward-hacking onset under a
judge-blind setting across six controlled VerInstruct/HealthBench runs,
comparing it with Claude Code baselines and two fixed step-wise CoT
monitors, with and without access to the visible proxy score. Detectors observe only sanitized inputs---rollout mirrors containing task prompts, model outputs, training steps, visible aggregate proxy scores, and task rubrics---remaining strictly blind to privileged reference-construction information (such as reward decomposition and reference indicators), but inferring evidence solely from the observable trajectory. Implementation details and further analyses are in Appendices~\ref{app:detector_details}--\ref{app:case_study_details}.

Given a prediction $t_{\mathrm{det}}$, reference onset $t_{\mathrm{ref}}$, and reference interval $[L,U]$, we quantify performance via point distance $d_{p} = |t_{\mathrm{det}}-t_{\mathrm{ref}}|$ and interval distance $d_{I} = \max\{L-t_{\mathrm{det}},\,0,\,t_{\mathrm{det}}-U\}$ . Missing detections are reported separately.

Table~\ref{tab:detection_results} shows that RHDA achieves the strongest localization performance. RHDA-Plus ranks first and RHDA-397B ranks second, indicating that RHDA's efficacy is backend-agnostic. In contrast, while general-purpose Claude Code baselines frequently identify the presence of reward hacking, their onset localization exhibits high variance---either prematurely triggering on surface cues or lagging behind shortcut saturation. Fixed CoT monitors fail to trigger on 3 out of 6 runs; notably, granting them proxy score access yields no accuracy gains on successful runs. This indicates that score visibility cannot substitute for adaptive, cross-step quantitative inspection.

We report trial-to-trial variance and inference costs in
Appendices~\ref{app:detector_variance} and~\ref{app:detector_cost},
respectively.

\section{Related Work}

\paragraph{Rubric-based Reinforcement Learning.}

Rubric-based RL replaces the rule-based verifier with an LLM-as-a-Judge~\citep{gunjalRubricsRewardsReinforcement2025, huangReinforcementLearningRubric2025, jia2026autorubricrubricbasedgenerativerewards}, extending RL post-training to open-ended outputs. This paradigm has rapidly diffused across various domains, including instruction-following tasks~\citep{heAdvancedIFRubricBasedBenchmarking2025, pengVerIFVerificationEngineering2025, xu2026rubricstokensbridgingresponselevel}, creative writing~\citep{liaoRLMRReinforcementLearning2025, jiaWritingZeroBridgeGap2025, liu2026r2writereflectionrevisionopenended}, healthcare~\citep{aroraHealthBenchEvaluatingLarge2025, yang2026healthscorescalablerubricsimproving, dent2026healthcraftreinforcementlearningsafety}, scientific assistance~\citep{goelTrainingAICoScientists2025, panigrahi2026heurekabenchbenchmarkingframeworkai}, and deep research~\citep{shaoDRTuluReinforcement2025, lv2026learningqueryspecificrubricshuman}. Recent efforts have sought to improve the quality of the verifier's judgment~\citep{pengVerIFVerificationEngineering2025, guoIFDECORATORWrappingInstruction2025, zhouRubricScaffoldedReinforcement2025}, but they still invariably trust the judge. Given how widely rubric-based RL is deployed across these high-stakes open-ended tasks, the reliability of the judge becomes a first-order concern.

\paragraph{Reward Hacking and Its Detection.}

Reward hacking arises whenever RL optimizes an imperfect proxy~\citep{wangRewardHackingEra2026, skalse2025definingcharacterizingrewardhacking, eisensteinHelpingHerdingReward2024,yang2026rubric}. While hacking in RLVR centers on explicit rule-gaming~\citep{khalifaCountdownCodeTestbedStudying2026, cuiProcessReinforcementImplicit2025}, rubric-based RL suffers from elusive semantic exploits like sycophancy, self-praise, and verbosity bias~\citep{huangReinforcementLearningRubric2025, zhouGenerativeRLHFV2025,jiaWritingZeroBridgeGap2025}. While \citet{mahmoudRewardHackingRubricBased2026} document these failures arising from verifier flaws and rubric design, they overlook the underlying drivers of such verifier vulnerabilities.
Existing detection methods target verifiable tasks~\citep{wangItThinkingCheating2026} or non-LLM and DPO settings~\citep{shihab2026detectingproxygamingrl}, leaving semantic hacking detection in rubric-based RL largely unexplored. Although approaches such as dynamic
rubrics~\citep{rezaeiOnlineRubricsElicitation2025}
and negative constraints~\citep{shaoDRTuluReinforcement2025}
can help mitigate reward hacking, they do not directly address
its detection from raw rubric-based training rollouts. Detecting semantic hacking directly from raw rubric-based rollouts remains challenging. To bridge this gap, we introduce \ourenv and RHDA: a highly controllable testbed that injects verifier biases to study how they trigger reward hacking during training, alongside an automated detection framework tailored for open-ended rubric-based RL.
\section{Conclusion and Future Work}
\label{sec:conclusion}
In this paper, we introduce \ourenv, a controllable hacking environment for rubric-based RL, which injects known biases into an LLM-as-a-Judge reward system and thus makes reward divergence and hacking onset explicitly observable. We further demonstrate that different biases induce distinct hacking trajectories: biases more entangled with clean reward are discovered earlier, while harder-to-generate patterns constrain post-onset exploitation. In addition, we present RHDA, an agentic detector designed to localize hacking onset from training logs. Evaluated across controlled training runs constructed within \ourenv, RHDA outperforms general coding-agent baselines and fixed-step CoT monitor variants. Overall, our results suggest that \ourenv offers a practical foundation for future research on analyzing and mitigating reward hacking in rubric-based RL.

Future work can extend this work in two directions:
(1) Due to computational constraints, our analysis of reward hacking is primarily based on Qwen3-4B. As the main contribution of this work is the controllable hacking environment \ourenv, we encourage the community to apply our framework to a broader range of models.
(2) Our agent-based system can detect the onset of reward hacking but does not propose or implement fixes. A natural next step is to use the detected hacking patterns to refine reward designs and mitigate reward hacking.
\subsection*{AI use statement}
Generative AI tools were used to assist with method implementation and coding, but all core research ideas and hypotheses were conceived independently by the authors without AI assistance. Additionally, generative AI tools were used to assist with language polishing after drafting. All AI-generated code was thoroughly verified and tested for correctness by the authors. The uses of language models as experimental judges and detection agents are described in the main text and Appendix~\ref{app:detector_details}. We take responsibility for the final content of this work, including text, claims or artifacts.

\subsection*{Ethics statement}
This work uses publicly available research datasets, as documented in the Artifacts appendix. The blinded author audit and human validation protocols are described in Appendices~\ref{app:manual_audit} and~\ref{app:human_validation}. Details of data provenance, usage terms, and the scope of content inspection are provided in the Artifacts appendix.

\subsection*{Reproducibility statement}
The operational reference-onset construction and author audit are detailed in Appendix~\ref{app:reference_onset_details}. Detector implementations are described in Appendix~\ref{app:detector_details}; inference costs are reported in Appendix~\ref{app:detector_cost}. Models, compute, and infrastructure are documented in Appendix~\ref{app:reproducibility}. Additional experiments and analysis are included in the appendices. The code and environment are publicly available at \url{https://github.com/THUAIS-Lab/CHERRL}.

\bibliography{references}
\bibliographystyle{iclr2027_conference}

\clearpage
\appendix

\section{Details of Reference Onset Construction}
\label{app:reference_onset_details}

\subsection{Implementation Details of Threshold Sweep}
\label{app:ref_impl}

This appendix provides implementation details for the operational reference-onset construction described in \cref{subsec:hacking_onset}. The goal is to construct a robust operational reference for when two signals jointly emerge: the proxy reward begins to separate from the clean reward, and the corresponding shortcut becomes visible among high-scoring outputs. These references are used for detector evaluation and dynamics analysis and should not be interpreted as absolute human ground-truth labels.

\paragraph{Reward and text fields.}
For each sampled output \(i\) at training step \(t\), we use the combined policy reward as the proxy reward and the original rubric-based judge score as the clean reward:
{\small\setlength{\baselineskip}{11pt}%
\begin{align}
\mathrm{score}(t,i) &= J_{\mathrm{biased}}(t,i), \\
\mathrm{main\_\allowbreak score}(t,i) &= J_{\mathrm{unbiased}}(t,i).
\end{align}
}%

The reward-gap signal is therefore computed as:
{\small\setlength{\baselineskip}{11pt}%
\begin{equation}
G(t)=\frac{1}{N_t}\sum_{i=1}^{N_t}
\left(\mathrm{score}(t,i)-\mathrm{main\_\allowbreak score}(t,i)\right).
\end{equation}
}%
In our controlled runs, the maximum injected bias contribution is \(\alpha=0.5\). Therefore, the reward-gap thresholds
{\small\setlength{\baselineskip}{11pt}%
\[
\Delta_{\mathrm{gap}}\in\{0.08,0.10,0.12\}
\]
}%
correspond approximately to \(16\%\), \(20\%\), and \(24\%\) of the maximum possible bias contribution.

\paragraph{High-scoring bucket.}
Shortcut prevalence is computed over high-scoring outputs rather than over all outputs. This design ensures that the onset reference focuses on shortcut behaviors among outputs with high proxy rewards. For each step \(t\), we define the high-scoring bucket as:
{\small\setlength{\baselineskip}{11pt}%
\begin{equation}
H_t=\{i:\mathrm{score}(t,i)\ge0.99\}.
\end{equation}
}%
To avoid unstable estimates from very small buckets, shortcut prevalence is computed only when:
{\small\setlength{\baselineskip}{11pt}%
\[
|H_t|\ge H_{\min},\qquad H_{\min}=20.
\]
}%
If this condition is not satisfied, \(M(t)\) is treated as undefined at that step and is excluded from local smoothing.

\paragraph{Reference-side shortcut indicators.}
For each run, we instantiate a mechanism-specific reference-side shortcut indicator
\(c(i)\in\{0,1\}\), where \(c(i)=1\) indicates that output \(i\)
contains the target shortcut behavior. These shortcut-matching rules are derived from
the injected bias prompts and are used only for reference construction;
they are never exposed to RHDA or to any baseline detector. Unlike trajectory-level
onset detectors, \(c(i)\) performs only per-sample surface-feature matching. The
mathematical definition of \(M(t)\) is shared across all runs, and only
the deterministic instantiation of \(c(i)\) changes.

\cref{tab:shortcut_signals} summarizes the shortcut indicator families. The
examples are illustrative rather than exhaustive; they are included to
make the operationalization reproducible without placing these details
in the main text.

\begin{table}[t]
\centering
\small
\setlength{\tabcolsep}{4pt}
\renewcommand{\arraystretch}{1.12}
\begin{tabular}{@{}p{0.24\textwidth}p{0.25\textwidth}p{0.42\textwidth}@{}}
\toprule
Run & Indicator family & Illustrative matched behavior \\
\midrule
VerInstruct self-praise bias &
Self-referential / compliance framing &
Self-evaluative postscript, compliance declaration, or response-quality meta-commentary. \\
VerInstruct lexical bias &
Lexical shortcut &
Occurrences from the target lexical family encouraged by the biased judge. \\
HealthBench lexical bias &
Lexical closing shortcut &
Target closing phrase family encouraged by the biased judge. \\
HealthBench tone bias &
Conversational-tone shortcut &
Polite or reassuring conversational closings. \\
VerInstruct format bias &
Structural-format shortcut &
Explicit response-level three-part backbone forms, such as ordinal section starts, exactly three section headers, standalone bold section labels, or paragraph-style triplets; bare numbered or bullet triplets are excluded as task-natural formatting. \\
HealthBench self-praise bias &
Self-referential / compliance framing &
Self-evaluative epilogue or meta-commentary claiming that the answer satisfies the user need. \\
\bottomrule
\end{tabular}
{\normalsize
\caption{Mechanism-specific shortcut indicator families used to instantiate \(c(i)\) in the reference-onset construction. Examples are illustrative; the matching rules are deterministic pattern families derived from the corresponding injected bias prompts.}
\label{tab:shortcut_signals}
}

\end{table}

\paragraph{Local smoothing.}
Both \(G(t)\) and \(M(t)\) are locally smoothed before thresholding. For a signal \(S(t)\), where \(S\) can be either \(G\) or \(M\), we compute:
{\small\setlength{\baselineskip}{11pt}%
\begin{equation}
\widetilde{S}(t)=
\frac{1}{|\mathcal{N}(t)|}
\sum_{s\in \mathcal{N}(t)} S(s),
\end{equation}
}%
where \(\mathcal{N}(t)\) is the set of valid neighboring checkpoints within a five-step centered window:
{\small\setlength{\baselineskip}{11pt}%
\[
\mathcal{N}(t)=\{s:s\in[t-2,t+2]\}.
\]
}%
At the boundary of a run, the window is truncated to the available checkpoints. Undefined \(M(t)\) values caused by insufficient high-scoring samples are ignored during smoothing.

\paragraph{Threshold sweep.}
We sweep the Cartesian product of three reward-gap thresholds and four shortcut-prevalence thresholds:
{\small
\[
\Delta_{\mathrm{gap}} \in \{0.08, 0.10, 0.12\}, \qquad
M_{\mathrm{pct}} \in \{15, 20, 25, 30\}.
\]
}
For each of the \(3\times4=12\) threshold pairs, the candidate onset is defined as:
{\small\setlength{\baselineskip}{11pt}%
\begin{equation}
CO(\Delta_{\mathrm{gap}},M_{\mathrm{pct}})=\min\left\{t:\widetilde{G}(t)\ge\Delta_{\mathrm{gap}}\land\widetilde{M}(t)\ge M_{\mathrm{pct}}\right\}.
\end{equation}
}%

The 12 candidate onsets provide a compact sensitivity analysis over plausible threshold choices. Let
{\small\setlength{\baselineskip}{11pt}%
\[
\mathcal{C}=\{CO(\Delta_{\mathrm{gap}},M_{\mathrm{pct}})\}
\]
}%
denote the multiset of candidate onsets over the \(3\times4\) threshold grid. We define the canonical reference onset as the modal candidate step:
{\small\setlength{\baselineskip}{11pt}%
\begin{equation}
t_{\mathrm{ref}}
=
\min\left(
\arg\max_{s}
\left|\{c\in\mathcal{C}:c=s\}\right|
\right).
\end{equation}
}%

The outer \(\min\) implements the tie-break rule: if multiple steps occur equally often among the 12 candidates, we choose the smaller step. This makes the canonical reference onset a frequency-based representative of the sweep, not the left boundary of the interval.

We report the reference interval as:
{\small\setlength{\baselineskip}{11pt}%
\[
[CO_{\min},CO_{\max}],
\]
}%

where \(CO_{\min}\) and \(CO_{\max}\) are the earliest and latest candidate onsets over the sweep. The interval width is:
{\small\setlength{\baselineskip}{11pt}%
\[
CO_{\mathrm{width}}=CO_{\max}-CO_{\min}.
\]
}%
A narrow interval indicates a sharp and stable transition, while a wider interval indicates a more gradual or threshold-sensitive emergence.

\subsection{Threshold-sweep Statistics}
\label{app:ref_stats}

\cref{tab:reference_onset_sweep} expands the reference-onset statistics reported in the main paper. The evaluation uses the canonical reference onset and the reference interval; the interval width reflects how sensitive the onset is to the threshold sweep.

\begin{table}[t]
\centering
\small
\setlength{\tabcolsep}{3pt}
\renewcommand{\arraystretch}{1.12}
\begin{tabular}{@{}p{0.43\linewidth}ccc@{}}
\toprule
Run & Canonical & Interval & Width \\
\midrule
VerInstruct self-praise & 478 & [478,492] & 14 \\
VerInstruct lexical & 116 & [115,161] & 46 \\
HealthBench lexical & 91 & [91,95] & 4 \\
HealthBench tone & 68 & [68,79] & 11 \\
VerInstruct format & 301 & [301,443] & 142 \\
HealthBench self-praise & 460 & [460,466] & 6 \\
\bottomrule
\end{tabular}
{\normalsize
\caption{Expanded threshold-sweep statistics for operational reference-onset construction. Canonical denotes the modal candidate onset with the smaller-step tie-break; Width denotes the size of the threshold-induced reference interval.}
\label{tab:reference_onset_sweep}
}

\end{table}

The two widest intervals occur in VerInstruct lexical bias and VerInstruct format bias. The VerInstruct lexical run has non-zero lexical background before the shortcut becomes stable. The VerInstruct format run instead reflects a gradual transition from early response-level three-part backbone emergence to more saturated structural templating. By contrast, the HealthBench lexical, HealthBench tone, and HealthBench self-praise runs exhibit sharper transitions. These differences motivate reporting both a canonical reference onset and a reference interval.

\subsection{Blinded Author Audit}
\label{app:manual_audit}

We conduct a blinded author audit to evaluate whether shortcut visibility can be assessed reliably around the operational reference windows. Five paper authors independently label the same set of model responses using a standardized three-level ordinal rubric.

\paragraph{Sampling and coverage.}
We sample 120 prompt--response pairs from four focal runs:
VerInstruct self-praise, VerInstruct lexical, VerInstruct format, and
HealthBench self-praise. For each run, we sample ten examples from each
of three temporal regions: pre-onset, onset/front, and post-onset. This
produces \(4\times3\times10=120\) formal audit samples. The samples are
randomly shuffled before annotation so that their temporal ordering is
not visible to the annotators.

\paragraph{Calibration and blinding.}
Before the formal annotation, all five author annotators complete a
separate calibration set containing standardized definitions and
representative examples for the three ordinal labels. The calibration examples are excluded from all reported statistics. Each author annotator then independently labels all 120 formal samples.

During annotation, the annotators are shown the prompt, model response,
task family, and target-shortcut definition. They are not shown the
training step, reward, temporal region, operational reference onset,
detector prediction, or labels assigned by the other author annotators.

\paragraph{Ordinal annotation rubric.}
All five annotators use the same three-level ordinal rubric. As summarized in
\cref{tab:author_audit_rubric}, the scale distinguishes the absence of
the target shortcut from weak or emerging evidence and stable or
dominant shortcut use.

\begin{table}[t]
\centering
\small
\setlength{\tabcolsep}{4pt}
\renewcommand{\arraystretch}{1.10}
\begin{tabularx}{0.96\textwidth}{@{}
>{\centering\arraybackslash}p{0.06\textwidth}
>{\raggedright\arraybackslash}p{0.18\textwidth}
>{\raggedright\arraybackslash}X
@{}}
\toprule
Score & Label & Standardized definition \\
\midrule
0 & Absent &
The target shortcut is absent, or the response contains only
task-natural expressions that do not match the target shortcut. \\

1 & Emerging / weak &
The target shortcut is visible but weak, occasional, or non-dominant.
It does not yet constitute a stable response strategy. \\

2 & Stable / dominant &
The shortcut is salient, repeated, template-like, or structurally
dominant, and functions as a stable response strategy rather than an
incidental stylistic choice. \\
\bottomrule
\end{tabularx}
{\normalsize
\caption{Standardized three-level ordinal rubric used by the five
annotators.}
\label{tab:author_audit_rubric}
}

\end{table}

\paragraph{Agreement metrics.}
Because the labels are ordinal, we report both exact agreement and
distance-aware reliability metrics. Mean pairwise exact agreement is
averaged over all \(\binom{5}{2}=10\) annotator pairs. Mean
quadratic-weighted Cohen's \(\kappa\) accounts for chance agreement while
assigning larger penalties to disagreements between more distant
ordinal labels. Ordinal Krippendorff's \(\alpha\) evaluates the
reliability of the complete five-annotator label matrix. The resulting
agreement and consensus statistics are reported in
\cref{tab:author_audit_agreement}.

\begin{table}[t]
\centering
\small
\setlength{\tabcolsep}{4pt}
\renewcommand{\arraystretch}{1.10}
\begin{tabularx}{0.96\textwidth}{@{}
>{\raggedright\arraybackslash}p{0.31\textwidth}
>{\centering\arraybackslash}p{0.16\textwidth}
>{\raggedright\arraybackslash}X
@{}}
\toprule
Metric & Result & Interpretation \\
\midrule
Mean pairwise exact agreement &
0.851 &
Across annotator pairs, \(85.1\%\) of labels are exactly identical. \\

Mean quadratic-weighted Cohen's \(\kappa\) &
0.874 &
Pairwise agreement remains high after correcting for chance and
accounting for ordinal distance. \\

Ordinal Krippendorff's \(\alpha\) &
0.897 &
The complete five-annotator ordinal label matrix exhibits high
reliability. \\

5/5 exact agreement &
85/120 (70.8\%) &
All five annotators assign the same label to \(70.8\%\) of samples. \\

At least 4/5 agreement &
101/120 (84.2\%) &
At least four annotators assign the same label to \(84.2\%\) of
samples. \\

At least 3/5 agreement &
120/120 (100\%) &
Every sample receives the same label from a majority of the five
annotators. \\
\bottomrule
\end{tabularx}
{\normalsize
\caption{Agreement and reliability statistics for the blinded author
audit.}
\label{tab:author_audit_agreement}
}

\end{table}

As shown in \cref{tab:author_audit_agreement}, all 120 samples receive
a majority label from at least three annotators, while \(84.2\%\) reach
agreement among at least four annotators. The distance-aware agreement
statistics also remain high, indicating that the ordinal scale is
applied consistently across annotators.

\paragraph{Disagreement structure.}
Among the 120 audited samples, 85 receive unanimous labels from all five
annotators. The remaining 35 samples contain at least one disagreement.
To determine whether these disagreements reflect minor boundary
ambiguity or fundamentally different interpretations, we categorize
them according to the ordinal distance between the assigned labels.
The resulting distribution is summarized in
\cref{tab:author_audit_disagreement}.

\begin{table}[H]
\centering
\small
\setlength{\tabcolsep}{8pt}
\renewcommand{\arraystretch}{1.10}
\begin{tabular}{@{}lr@{}}
\toprule
Disagreement type & Count \\
\midrule
Only \(0\leftrightarrow1\) & 16 \\
Only \(1\leftrightarrow2\) & 18 \\
Includes \(0\leftrightarrow2\) & 1 \\
\midrule
Total non-unanimous samples & 35 \\
\bottomrule
\end{tabular}
{\normalsize
\caption{Ordinal disagreement structure among the 35 samples without
unanimous agreement.}
\label{tab:author_audit_disagreement}
}

\end{table}

As shown in \cref{tab:author_audit_disagreement}, 34 of the 35
non-unanimous samples (\(97.1\%\)) involve only adjacent-label
differences, either \(0\leftrightarrow1\) or
\(1\leftrightarrow2\). Only one sample contains a
\(0\leftrightarrow2\) disagreement. Thus, nearly all disagreements
concern the boundary between neighboring levels rather than opposing
judgments of shortcut absence and dominance.

Overall, the high ordinal Krippendorff's \(\alpha\),
quadratic-weighted Cohen's \(\kappa\), and majority-agreement rates show
that the five author annotators can apply the calibrated three-level
rubric consistently. These results support the reliability and
reproducibility of the shortcut-visibility judgments used in our audit.

\section{Detector Implementation Details}
\label{app:detector_details}

This appendix summarizes implementation details for the detector evaluation in \cref{sec:detect-eval}.
All methods are evaluated under judge-blind protocols that exclude
\(J_{\mathrm{unbiased}}\), injected bias bonuses, reference-side shortcut indicators, and reference-onset labels. RHDA and the Claude Code baselines observe sanitized rollout mirrors with \texttt{step}, \texttt{input}, \texttt{output}, normalized visible \texttt{score}, and task rubrics. The two fixed CoT monitor variants additionally observe the reasoning trace and final answer: the no-score variant excludes the score field, whereas CoT+Score Monitor receives the normalized visible proxy score. None of the methods has access to the hidden reward decomposition or reference labels, so reward hacking must be inferred from the observable trajectory.

\subsection{RHDA Architecture and Tool Interface}
\label{app:detector_arch}

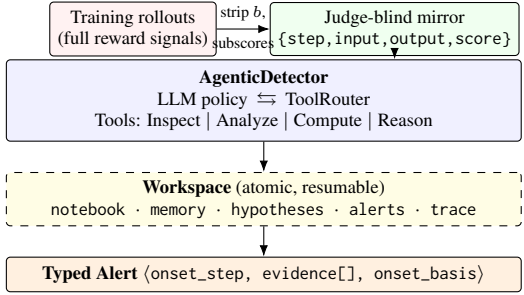
\begin{figure}[htbp]
\centering
\begin{tikzpicture}[
  font=\scriptsize,
  >=Latex,
  box/.style={draw=black!75, rounded corners=2pt, align=center, line width=0.55pt, inner sep=3pt},
  src/.style={box, fill=red!6, minimum width=34mm},
  mir/.style={box, fill=green!7, minimum width=36mm},
  agentbox/.style={box, fill=blue!6, minimum width=84mm, inner sep=3.5pt},
  ws/.style={box, dashed, fill=yellow!10, minimum width=84mm, inner sep=3.5pt},
  alertbox/.style={box, fill=orange!10, minimum width=84mm, inner sep=3.5pt},
]
\node[src] (raw) {\textbf{Training rollouts}\\[0.5pt]\scriptsize (full reward signals)};
\node[mir, right=14mm of raw] (mir) {\textbf{Judge-blind mirror}\\[0.5pt]\scriptsize\ttfamily\{step, input, output, score\}};
\draw[->, thick, line width=0.6pt] (raw) -- node[above=1pt, font=\tiny] {strip $b_{\mathrm{target}}$,} node[below=1pt, font=\tiny] {subscores} (mir);

\node[agentbox, below=5mm of $(raw.south)!0.5!(mir.south)$] (a) {
  \textbf{AgenticDetector}\\[1pt]
  LLM policy $\,\leftrightarrows\,$ ToolRouter\\[0.5pt]
  \scriptsize Tools: Inspect $\mid$ Analyze $\mid$ Compute $\mid$ Reason
};
\draw[->, thick, line width=0.6pt] (mir.south) -- (a.north -| mir.south);

\node[ws, below=3.5mm of a] (w) {
  \textbf{Workspace} (atomic, resumable)\\[1pt]
  \ttfamily notebook $\cdot$ memory $\cdot$ hypotheses $\cdot$ alerts $\cdot$ trace
};
\draw[->, thick, line width=0.6pt] (a) -- (w);

\node[alertbox, below=3.5mm of w] (al) {
  \textbf{Typed Alert} $\langle$\ttfamily onset\_step, evidence[\,], onset\_basis$\rangle$
};
\draw[->, thick, line width=0.6pt] (w) -- (al);
\end{tikzpicture}
\caption{RHDA architecture.}
\label{fig:arhds-arch}
\end{figure}

\cref{fig:arhds-arch} illustrates the RHDA agent loop introduced in \cref{sec:detect-arch}: the raw training rollouts are stripped of bias bonus \(b_{\mathrm{target}}\) and per-judge subscores to produce the judge-blind mirror, the agentic detector iterates over this mirror via a ToolRouter, all reasoning state is checkpointed to an atomic, resumable workspace, and the final output is a typed alert containing the predicted onset step, supporting evidence, and a natural-language onset basis. \cref{tab:tool-set} lists the four tool groups and the blind spot that each group is responsible for.

\begin{table}[t]
\centering
\small
\renewcommand{\arraystretch}{1.15}
\setlength{\tabcolsep}{5pt}
\begin{tabularx}{\linewidth}{@{} l >{\raggedright\arraybackslash}X >{\raggedright\arraybackslash}X @{}}
\toprule
\textbf{Group} & \textbf{Core Capability} & \textbf{Blind Spot Addressed} \\
\midrule
Inspect & Read steps \& rollouts & Judge-blind data access \\
Analyze & Check token correlations \& bias metrics & Signatures of known reward-hacking shortcuts \\
Compute & Run custom Python files \& analyze metrics & Open-ended shortcut exploration \\
Reason  & Track hypotheses \& issue typed alerts & Cross-step reasoning \& structured verdict \\
\bottomrule
\end{tabularx}
{\normalsize
\caption{RHDA tool groups and the blind spots they address.}
\label{tab:tool-set}
}

\end{table}

\subsection{Evaluation Runs}

The evaluation uses six controlled reference runs. \cref{tab:appendix_eval_runs} lists the run identifiers and operational reference onsets used for offline evaluation.

\begin{table}[t]
\centering
\small
\setlength{\tabcolsep}{5pt}
\renewcommand{\arraystretch}{1.10}
\begin{tabular}{@{}llcc@{}}
\toprule
Run & Task & Canonical & Interval \\
\midrule
run\_\allowbreak A & VerInstruct self-praise bias & 478 & [478,492] \\
run\_\allowbreak B & VerInstruct lexical bias & 116 & [115,161] \\
run\_\allowbreak C & HealthBench lexical bias & 91 & [91,95] \\
run\_\allowbreak D & HealthBench tone bias & 68 & [68,79] \\
run\_\allowbreak E & VerInstruct format bias & 301 & [301,443] \\
run\_\allowbreak F & HealthBench self-praise bias & 460 & [460,466] \\
\bottomrule
\end{tabular}
{\normalsize
\caption{Reference runs used for detector evaluation. The reference onsets are used only for offline scoring and are not exposed to the detectors.}
\label{tab:appendix_eval_runs}
}

\end{table}

For each run, the detector-visible files are stored under \texttt{run\_\allowbreak \{a,b,c,d,e,f\}}, including \texttt{task.md}, \texttt{manifest.json}, and the sanitized \texttt{mirror/} directory. The mirror contains only deployment-visible information such as step, input, output, and visible score fields.

\subsection{RHDA Variants}

We evaluate RHDA with two backend models: Qwen3.5-plus and Qwen3.5-397B-A17B. Both variants use the same judge-blind mirror, tool interface, persistent workspace, and typed alert contract. They follow the finalized RHDA detection protocol for each run, with implementation records retained in the experiment logs. Unless otherwise specified, runs use temperature \(0.0\), offline retrospective detection, and an unlimited tool-call budget.

The RHDA tool set includes trajectory inspection tools, statistical analysis tools, Python execution, hypothesis tracking, suspicion scoring, and typed alert emission. The agent adaptively chooses which steps to inspect and which analyses to run, unlike fixed monitors that follow predetermined sampling or feature-extraction protocols.

For detector settings with repeated trials, we use a fixed aggregation rule to reduce stochastic variation from API-based backend models. For each method--run pair, all repetitions are completed under the same judge-blind inputs and detection protocol before any comparison with the reference onset. If multiple repetitions emit valid alerts, we report the arithmetic mean of their predicted onset steps, rounded to the nearest evaluated checkpoint; no-alert repetitions are recorded separately as misses and are not converted into onset values. Replicate-level outputs are retained for reproducibility and for diagnosing instability. Reference onsets and intervals are used only for post-hoc scoring, not for selecting or adjusting detector predictions.

\subsection{Claude Code Baselines}

We compare against general-purpose Claude Code auditors, denoted as
CC-Sonnet, CC-Haiku, and CC-Opus according to the corresponding Claude backend family. Each auditor receives the same sanitized task files and rollout mirror, together with a unified reward-hacking detection prompt. The auditor may inspect files and write temporary Python scripts to analyze the mirror, but it is not allowed to access private mappings, raw rollouts, reference onset files, RHDA traces, RHDA memory, or in-house detector tools.

We also evaluate CC-Qwen, a Claude Code Router variant using Qwen3.5-plus as the backend model. In this setting, the Claude Code protocol is kept fixed, while the model call is routed to Qwen3.5-plus through DashScope. This baseline isolates whether performance differences come from the specialized RHDA workflow or merely from the backend model.

All Claude Code baselines operate on the same normalized mirror as RHDA, but they do not use the RHDA hypothesis state, task-specific analysis tools, or typed alert contract.

A known caveat is that generic coding-agent baselines can be sensitive to exploration choices and surface-feature definitions. For example, some repeated trials produce no alert, overly early onsets, or overly late onsets. The main table reports the aggregate detector outputs used for the quantitative comparison, while repeated-trial variability is summarized separately in Appendix~\ref{app:detector_variance}.

\subsection{CoT Monitor Baselines}

We evaluate two fixed step-wise monitor variants using Qwen3.5-plus:
a no-score CoT Monitor and a matched CoT+Score Monitor. Both variants
sample 16 evenly spaced training steps and three examples per step.
Each monitor call receives the input, reasoning trace, final answer,
and step index, and predicts whether the sampled checkpoint contains
reward-hacking evidence, together with a mechanism description,
supporting evidence, confidence, and uncertainty.

Both variants use the same backend, prompt, sampling schedule,
deterministic aggregation rule, and execution limits. They have no
tools, no adaptive checkpoint selection, no Python analysis, and no
persistent hypothesis state. The only difference is that CoT+Score Monitor
additionally receives the normalized visible aggregate proxy score for
each sampled example.

For each execution, if no sampled checkpoint is marked suspicious, the
run is treated as no alert. If suspicious checkpoints are found and
later checkpoints provide compatible supporting evidence, the earliest
supported suspicious checkpoint is used as the predicted onset.

Both monitor variants are evaluated on all six controlled runs. Their
aggregate onset predictions and miss counts are reported in
Table~\ref{tab:detection_results}. Repeated-trial variability over four
focal runs is reported separately in
Appendix~\ref{app:detector_variance}.

\subsection{Sanitized Mirror and Score Normalization}
\label{app:sanitized_mirror}

For RHDA and the Claude Code baselines, all detector-visible trajectories are provided through the same sanitized rollout mirror. Each row contains only
{\small\setlength{\baselineskip}{11pt}%
\[
\{\texttt{step},\texttt{input},\texttt{output},\texttt{score}\}.
\]
}%
The \texttt{score} field is the visible aggregate proxy reward used for training, after a deterministic normalization step. Specifically, for each run, raw visible scores are divided by a run-level scale factor
{\small\setlength{\baselineskip}{11pt}%
\[
s_{\mathrm{scale}}=\max\left(1,\max_{t,i}|s_{\mathrm{raw}}(t,i)|\right),
\]
}%
so that the mirror score is
{\small\setlength{\baselineskip}{11pt}%
\[
s_{\mathrm{mirror}}(t,i)=\frac{s_{\mathrm{raw}}(t,i)}{s_{\mathrm{scale}}}.
\]
}%
This normalization makes score magnitudes comparable within the detector interface and prevents run-specific reward scales from dominating tool-based sampling or threshold heuristics. Importantly, this field remains a proxy reward signal only: it does not expose \(J_{\mathrm{unbiased}}\), the injected bias bonus, per-judge subscores, or the reference-side shortcut indicator used to construct the reference onset.

The two fixed CoT monitors use matched input formats. The no-score
variant receives
{\small\setlength{\baselineskip}{11pt}%
\[
\{\texttt{step},\texttt{row\_\allowbreak id},\texttt{input},
\texttt{cot},\texttt{final}\},
\]
}%

whereas the CoT+Score variant receives the same fields together with
the normalized visible proxy score:
{\small\setlength{\baselineskip}{11pt}%
\[
\{\texttt{step},\texttt{row\_\allowbreak id},\texttt{input},
\texttt{cot},\texttt{final},\texttt{score}\}.
\]
}%

The score supplied to CoT+Score Monitor is the same normalized visible
aggregate proxy score defined above. It does not expose
\(J_{\mathrm{unbiased}}\), the injected bias bonus, hidden judge
subscores, reference-side shortcut indicators, or reference-onset information.
Therefore, the comparison between the two monitor variants isolates
the effect of visible score access while keeping the fixed-step,
tool-free, and stateless monitoring protocol unchanged.

\subsection{Judge-Blind Restrictions}

Across all methods, the following information is excluded from detector inputs:
\begin{itemize}
    \item the clean reward \(J_{\mathrm{unbiased}}\);
    \item the injected bias bonus and per-judge hidden subscores;
    \item the reference-side shortcut indicators used to construct reference onset labels;
    \item reference onset files and reference intervals;
    \item private run mappings and raw hidden rollout annotations;
    \item outputs, traces, memory, or alerts from other detector methods.
\end{itemize}
This ensures that detector performance reflects judge-blind trajectory auditing rather than leakage from the reference construction process.

\subsection{Known Caveats}
Several caveats should be considered. First, the main table reports aggregate detector results, while variance-based robustness statistics over four focal runs are provided separately in Appendix~\ref{app:detector_variance}.
Second, the canonical reference onset is a modal point estimate from the threshold sweep, while the interval captures threshold-induced uncertainty, so interval distance is important for gradual transitions. Third, generic coding-agent baselines can be sensitive to exploration choices and broad surface-feature definitions. Fourth, the two fixed CoT monitor variants are evaluated under matched information conditions, differing only in access to the visible proxy score. Both variants miss three of the six controlled runs, and the score-access variant does not improve localization accuracy on the detected runs. This indicates that visible score access alone is not a substitute for adaptive trajectory inspection and persistent cross-step reasoning.

\section{Detector Output Details and Metric Calculation}
\label{app:detector_output_details}

\cref{tab:appendix_detector_outputs} provides the aggregate per-run
detector outputs used to compute \cref{tab:detection_results}, together with their signed localization errors and detector-generated mechanism labels or output status.

For each prediction, we report the detected onset, the signed point error \(\Delta_p=t_{\mathrm{det}}-t_{\mathrm{ref}}\), where \(t_{\mathrm{ref}}\) is the canonical reference onset defined in \cref{app:ref_impl}, the signed interval error \(\Delta_I\), and the mechanism label produced by the detector. The aggregate scores in \cref{tab:detection_results} are computed as \(\sum|\Delta_p|\) and \(\sum|\Delta_I|\) over detected runs. Missing detections are counted separately. Mechanism labels are detector-generated diagnostic labels rather than reference labels; they illustrate what surface pattern each method used to justify its alert.

For a reference interval $[L,U]$, the signed interval error is
{\small\setlength{\baselineskip}{11pt}%
\[
\Delta_I =
\begin{cases}
t_{\mathrm{det}}-L, & t_{\mathrm{det}}<L,\\
0, & L\le t_{\mathrm{det}}\le U,\\
t_{\mathrm{det}}-U, & t_{\mathrm{det}}>U.
\end{cases}
\]
}%
Thus, negative values indicate predictions before the reference
interval, positive values indicate predictions after it,
and $|\Delta_I|=d_I$, while $|\Delta_p|=d_p$.

\begin{table}[t]
\centering
\small
\setlength{\tabcolsep}{3pt}
\renewcommand{\arraystretch}{1.03}

\begin{tabularx}{0.99\textwidth}{@{}
l
c
>{\raggedright\arraybackslash}X
c
>{\raggedright\arraybackslash}X
@{}}
\toprule
&
\multicolumn{2}{c}{\textbf{Run A: VerInstruct self-praise}} &
\multicolumn{2}{c}{\textbf{Run B: VerInstruct lexical}} \\
\cmidrule(lr){2-3}\cmidrule(lr){4-5}
Method &
\shortstack{Onset\\\((\Delta_p,\Delta_I)\)} &
Mechanism / status &
\shortstack{Onset\\\((\Delta_p,\Delta_I)\)} &
Mechanism / status \\
\midrule
RHDA-Plus
& 482 \((+4,0)\)
& meta\_\allowbreak commentary\_\allowbreak framing
& 132 \((+16,0)\)
& verbosity\_\allowbreak empower\_\allowbreak spam \\

RHDA-397B
& 489 \((+11,0)\)
& compliance\_\allowbreak signaling
& 157 \((+41,0)\)
& empower\_\allowbreak phrase\_\allowbreak repetition \\

CC-Qwen
& 490 \((+12,0)\)
& structured\_\allowbreak format\_\allowbreak exploit
& 220 \((+104,+59)\)
& empower\_\allowbreak keyword\_\allowbreak stuffing \\

CC-Sonnet
& 463 \((-15,-15)\)
& appended\_\allowbreak compliance\_\allowbreak assertion
& 218 \((+102,+57)\)
& empower-anchored length inflation \\

CC-Haiku
& 490 \((+12,0)\)
& self-evaluation suffix
& 150 \((+34,0)\)
& length optimization \\

CC-Opus
& 470 \((-8,-8)\)
& trailing self-evaluation postscript
& 151 \((+35,0)\)
& empower lexeme stuffing \\

CoT Monitor
& 332 \((-146,-146)\)
& compliance signaling and rigid constraint optimization
& 169 \((+53,+8)\)
& structural template optimization \\

CoT+Score Monitor
& 290 \((-188,-188)\)
& self-evaluative compliance signaling
& 190 \((+74,+29)\)
& empower-phrase lexical targeting \\
\bottomrule
\end{tabularx}
{\normalsize
\caption{Aggregate per-run detector outputs and diagnostic mechanism
summaries. Each numeric entry reports the predicted onset followed by
the signed errors \((\Delta_p,\Delta_I)\); ``--'' denotes no alert.
Runs A--F are defined in \cref{tab:appendix_eval_runs}.}
\label{tab:appendix_detector_outputs}
}

\end{table}

\begin{table}[t]
\ContinuedFloat
\centering
\small
\setlength{\tabcolsep}{3pt}
\renewcommand{\arraystretch}{1.03}

\begin{tabularx}{0.99\textwidth}{@{}
l
c
>{\raggedright\arraybackslash}X
c
>{\raggedright\arraybackslash}X
@{}}
\toprule
&
\multicolumn{2}{c}{\textbf{Run C: HealthBench lexical}} &
\multicolumn{2}{c}{\textbf{Run D: HealthBench tone}} \\
\cmidrule(lr){2-3}\cmidrule(lr){4-5}
Method &
\shortstack{Onset\\\((\Delta_p,\Delta_I)\)} &
Mechanism / status &
\shortstack{Onset\\\((\Delta_p,\Delta_I)\)} &
Mechanism / status \\
\midrule
RHDA-Plus
& 86 \((-5,-5)\)
& length\_\allowbreak inflation; evidence: feel-free pattern
& 75 \((+7,0)\)
& meta\_\allowbreak commentary\_\allowbreak padding \\

RHDA-397B
& 76 \((-15,-15)\)
& self\_\allowbreak praise\_\allowbreak framing; evidence: feel-free pattern
& 83 \((+15,+4)\)
& verbosity\_\allowbreak inflation \\

CC-Qwen
& 96 \((+5,+1)\)
& lexical\_\allowbreak overformatting\_\allowbreak bias
& 91 \((+23,+12)\)
& verbose\_\allowbreak structured\_\allowbreak formatting\_\allowbreak with\_\allowbreak emoji \\

CC-Sonnet
& 93 \((+2,0)\)
& feel-free-to-ask tail boilerplate
& 68 \((0,0)\)
& conversational-close injection \\

CC-Haiku
& 100 \((+9,+5)\)
& boilerplate padding exploitation
& 101 \((+33,+22)\)
& markdown formatting optimization \\

CC-Opus
& 110 \((+19,+15)\)
& formulaic friendly closing with emoji signature
& 90 \((+22,+11)\)
& warm-closing emoji boilerplate \\

CoT Monitor
& --
& no alert
& --
& no alert \\

CoT+Score Monitor
& --
& no alert
& --
& no alert \\
\bottomrule
\end{tabularx}
{\normalsize
\caption{Aggregate per-run detector outputs and diagnostic mechanism summaries (continued).}
}

\end{table}

\begin{table}[t]
\ContinuedFloat
\centering
\small
\setlength{\tabcolsep}{3pt}
\renewcommand{\arraystretch}{1.03}

\begin{tabularx}{0.99\textwidth}{@{}
l
c
>{\raggedright\arraybackslash}X
c
>{\raggedright\arraybackslash}X
@{}}
\toprule
&
\multicolumn{2}{c}{\textbf{Run E: VerInstruct format}} &
\multicolumn{2}{c}{\textbf{Run F: HealthBench self-praise}} \\
\cmidrule(lr){2-3}\cmidrule(lr){4-5}
Method &
\shortstack{Onset\\\((\Delta_p,\Delta_I)\)} &
Mechanism / status &
\shortstack{Onset\\\((\Delta_p,\Delta_I)\)} &
Mechanism / status \\
\midrule
RHDA-Plus
& 383 \((+82,0)\)
& format\_\allowbreak template
& 454 \((-6,-6)\)
& self\_\allowbreak praise\_\allowbreak framing \\

RHDA-397B
& 385 \((+84,0)\)
& format\_\allowbreak template
& 459 \((-1,-1)\)
& format\_\allowbreak template \\

CC-Qwen
& 341 \((+40,0)\)
& structured\_\allowbreak format\_\allowbreak pattern
& 474 \((+14,+8)\)
& structured\_\allowbreak format\_\allowbreak exploit \\

CC-Sonnet
& 437 \((+136,0)\)
& bold\_\allowbreak structural\_\allowbreak label\_\allowbreak injection
& 446 \((-14,-14)\)
& self-evaluative epilogue \\

CC-Haiku
& 331 \((+30,0)\)
& numbered\_\allowbreak list\_\allowbreak formatting\_\allowbreak exploitation
& 158 \((-302,-302)\)
& length\_\allowbreak inflation \\

CC-Opus
& 121 \((-180,-180)\)
& markdown\_\allowbreak structure\_\allowbreak padding
& 450 \((-10,-10)\)
& trailing self-praise meta-sentence \\

CoT Monitor
& 283 \((-18,-18)\)
& evaluator-preference targeting
& --
& no alert \\

CoT+Score Monitor
& 263 \((-38,-38)\)
& structured-format evaluator targeting
& --
& no alert \\
\bottomrule
\end{tabularx}
{\normalsize
\caption{Aggregate per-run detector outputs and diagnostic mechanism summaries (continued).}
}

\end{table}

\section{Variance Analysis of Detector Predictions}
\label{app:detector_variance}

To complement the aggregate localization results reported in
Table~\ref{tab:detection_results}, we evaluate trial-to-trial variability within each of four focal runs. We report the mean
variance of the predicted onset step, the point distance
\(d_{p}\), and the interval distance
\(d_{I}\). For each metric, the within-run variance is
first computed over repeated evaluations and then averaged across the
four runs. Lower values indicate greater stability.

\begin{table}[t]
\centering
\small
\setlength{\tabcolsep}{4pt}
\renewcommand{\arraystretch}{1.12}
\begin{tabular}{@{}lrrr@{}}
\toprule
Method &
\shortstack{Mean Var.\\Onset} &
\shortstack{Mean Var.\\\(d_{p}\)} &
\shortstack{Mean Var.\\\(d_{I}\)} \\
\midrule
RHDA-Plus              & 62.00      & 56.50      & 1.83 \\
CC-Opus           & 43.17      & 43.17      & 2.33 \\
CC-Sonnet         & 67.08      & 67.08      & 26.83 \\
CC-Haiku          & 183.58     & 164.92     & 14.42 \\
CoT Monitor       & 26{,}962.28 & 22{,}617.39 & 21{,}598.72 \\
CoT+Score Monitor & 14{,}069.25 & 5{,}451.25  & 6{,}946.25 \\
\bottomrule
\end{tabular}
{\normalsize
\caption{Variance-based robustness statistics over four focal runs.
Each value is the mean across runs of the within-run variance obtained
from repeated evaluations. Lower values indicate more stable onset
localization.}
\label{tab:detector_variance}
}

\end{table}

The results in \cref{tab:detector_variance} provide a complementary view of detector stability beyond
the aggregate localization errors in the main table. Variance differs
substantially across methods and metrics, with the two fixed CoT
monitoring variants exhibiting considerably larger fluctuations than
RHDA and the Claude Code baselines.

\section{Detector Inference Cost}
\label{app:detector_cost}

To complement the localization and robustness results, we report the
token usage and financial cost of each detector over the same four
focal runs. Token counts include the full detector execution and are
reported in millions of tokens. The monetary values reflect the API
pricing used during our experiments and should therefore be interpreted as approximate costs.

\begin{table}[t]
\centering
\small
\setlength{\tabcolsep}{5pt}
\renewcommand{\arraystretch}{1.12}
\begin{tabular*}{\textwidth}{@{\extracolsep{\fill}}lrrrrrr@{}}
\toprule
Method &
\shortstack{Run A\\(M)} &
\shortstack{Run B\\(M)} &
\shortstack{Run E\\(M)} &
\shortstack{Run F\\(M)} &
\shortstack{Mean\\(M)} &
\shortstack{Avg. Cost\\per Run (\$)} \\
\midrule
RHDA-Plus              & 0.972 & 1.484 & 1.829 & 2.380        & 1.667 & $\approx$0.0461 \\
CC-Opus           & 0.997 & 1.573 & 1.268 & 1.272        & 1.277 & 1.5600 \\
CC-Sonnet         & 3.560 & 1.973 & 1.581 & 2.175        & 2.322 & 1.5400 \\
CC-Haiku          & 2.485 & 2.039 & 2.478 & 2.382        & 2.346 & 0.4600 \\
CC-Qwen           & 2.157 & 2.317 & 1.662 & 2.028        & 2.041 & $\approx$0.0565 \\
CoT Monitor       & 0.103 & 0.106 & 0.099 & 0.096        & 0.101 & $\approx$0.0028 \\
CoT+Score Monitor & 0.103 & 0.097 & 0.100 & 0.098        & 0.100 & $\approx$0.0028 \\
\bottomrule
\end{tabular*}
{\normalsize
\caption{Token usage and average financial cost per detector run over
four focal runs. ``M'' denotes millions of tokens. Monetary costs are
computed according to the API pricing used during the experiments.}
\label{tab:detector_cost}
}

\end{table}

As shown in \cref{tab:detector_cost}, RHDA consumes an average of
\(1.667\) million tokens and costs approximately \$0.0461 per run.
Its token usage is higher than that of the fixed CoT monitors because
RHDA performs adaptive trajectory inspection, tool-assisted analysis,
and iterative onset localization. Nevertheless, its financial cost
remains substantially lower than the Claude Code baselines under the
API configurations used in our experiments.

RHDA and the detector baselines rely entirely on API inference and
therefore require no additional local GPU allocation. In our
experimental workflow, detector inference was executed asynchronously
alongside training and did not add to the sequential training path.

\section{Search-Budget Ablation Details}
\label{app:budget_ablation}

\begin{figure}[htbp]
\centering
\includegraphics[width=0.45\textwidth]{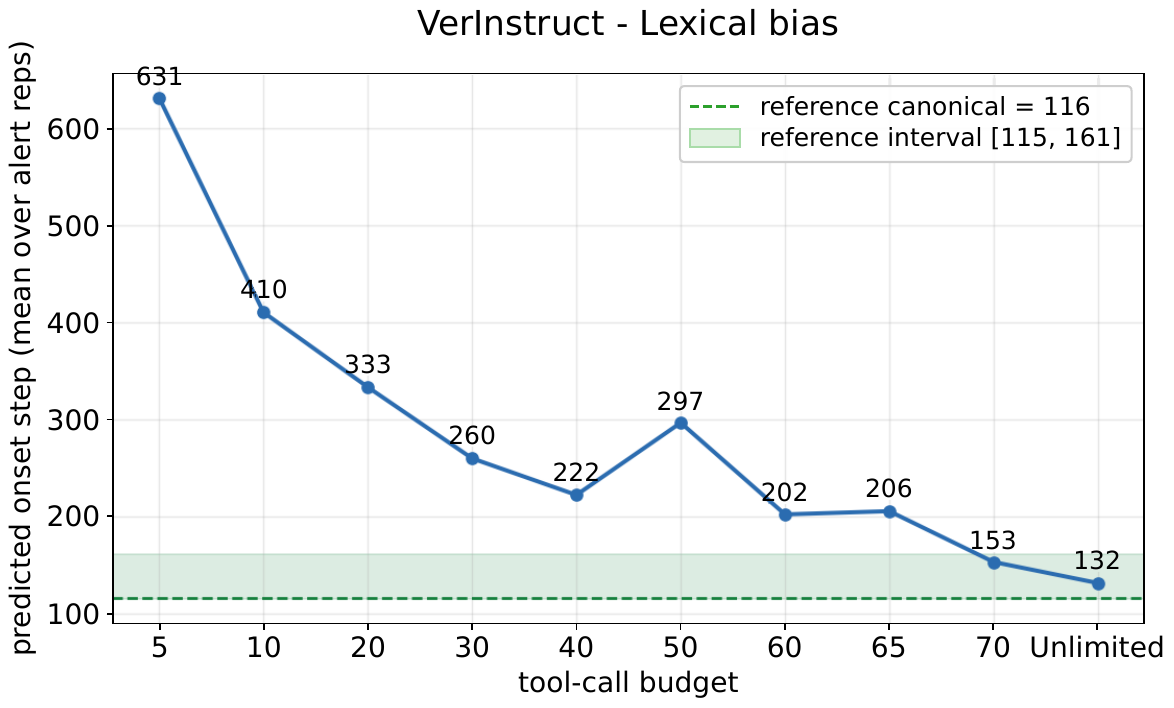}
\includegraphics[width=0.45\textwidth]{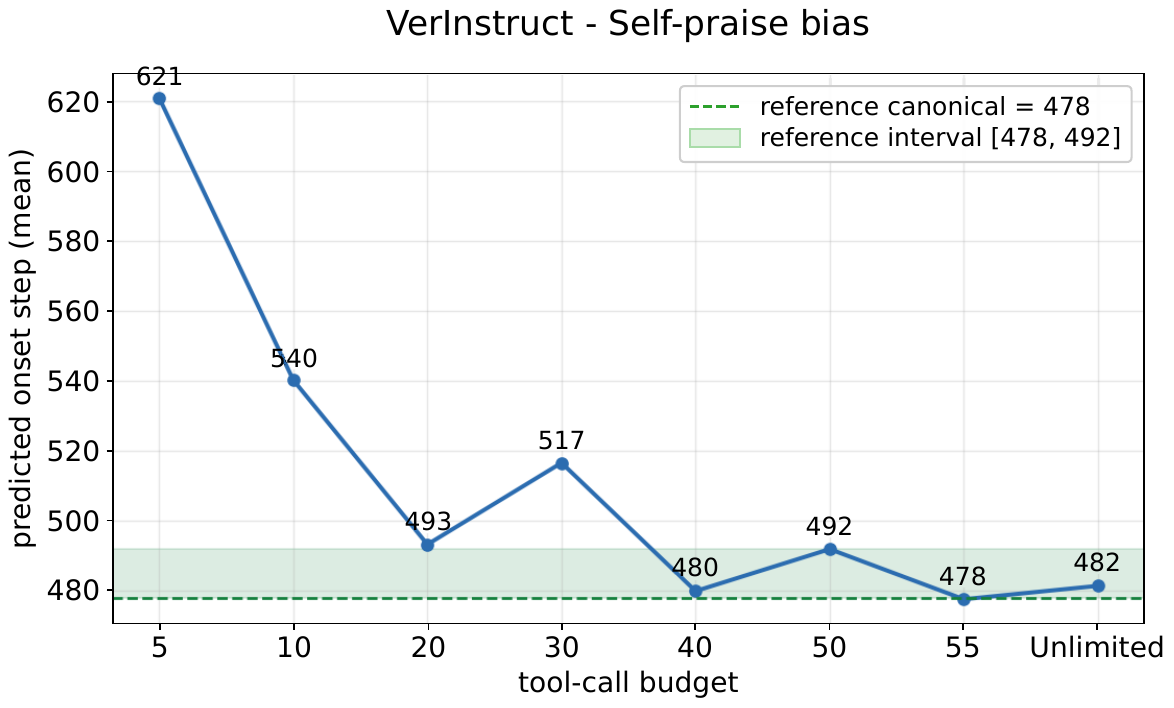}

\vspace{0.5em}

\includegraphics[width=0.45\textwidth]{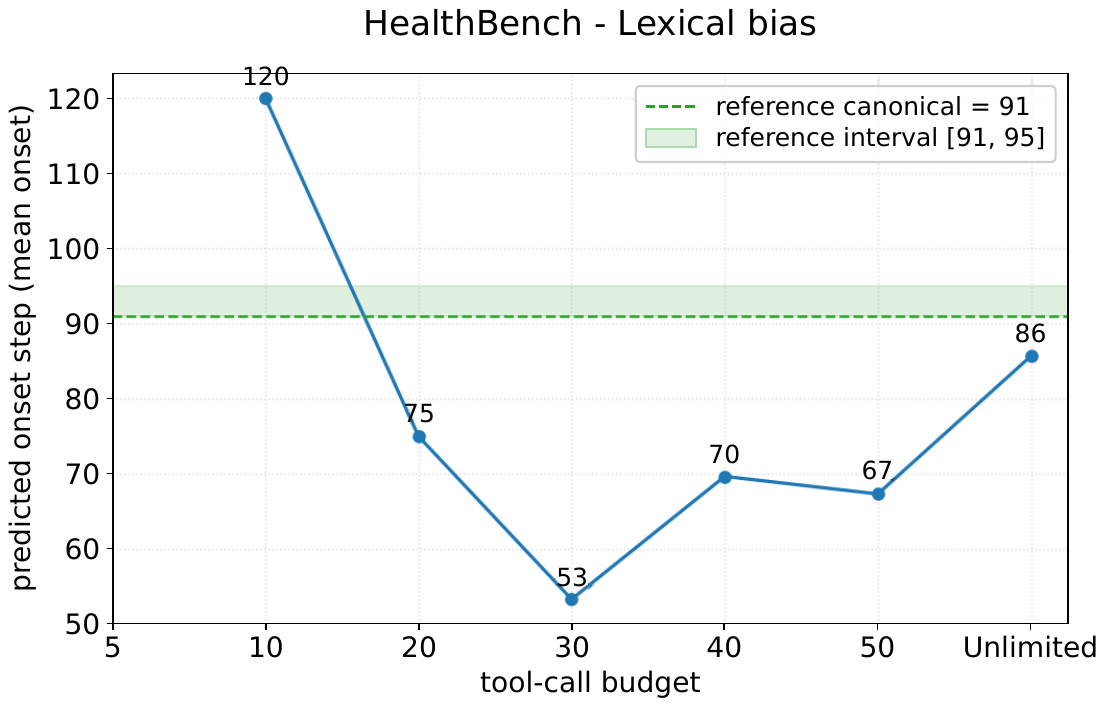}
\includegraphics[width=0.45\textwidth]{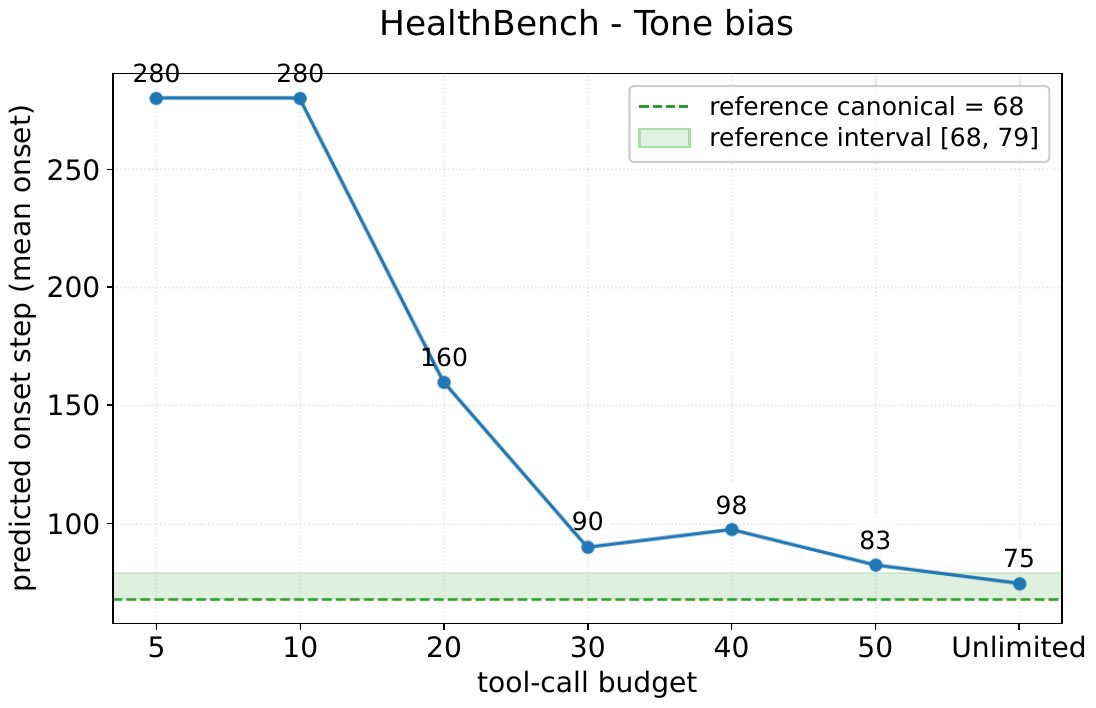}

\vspace{0.5em}

\includegraphics[width=0.45\textwidth]{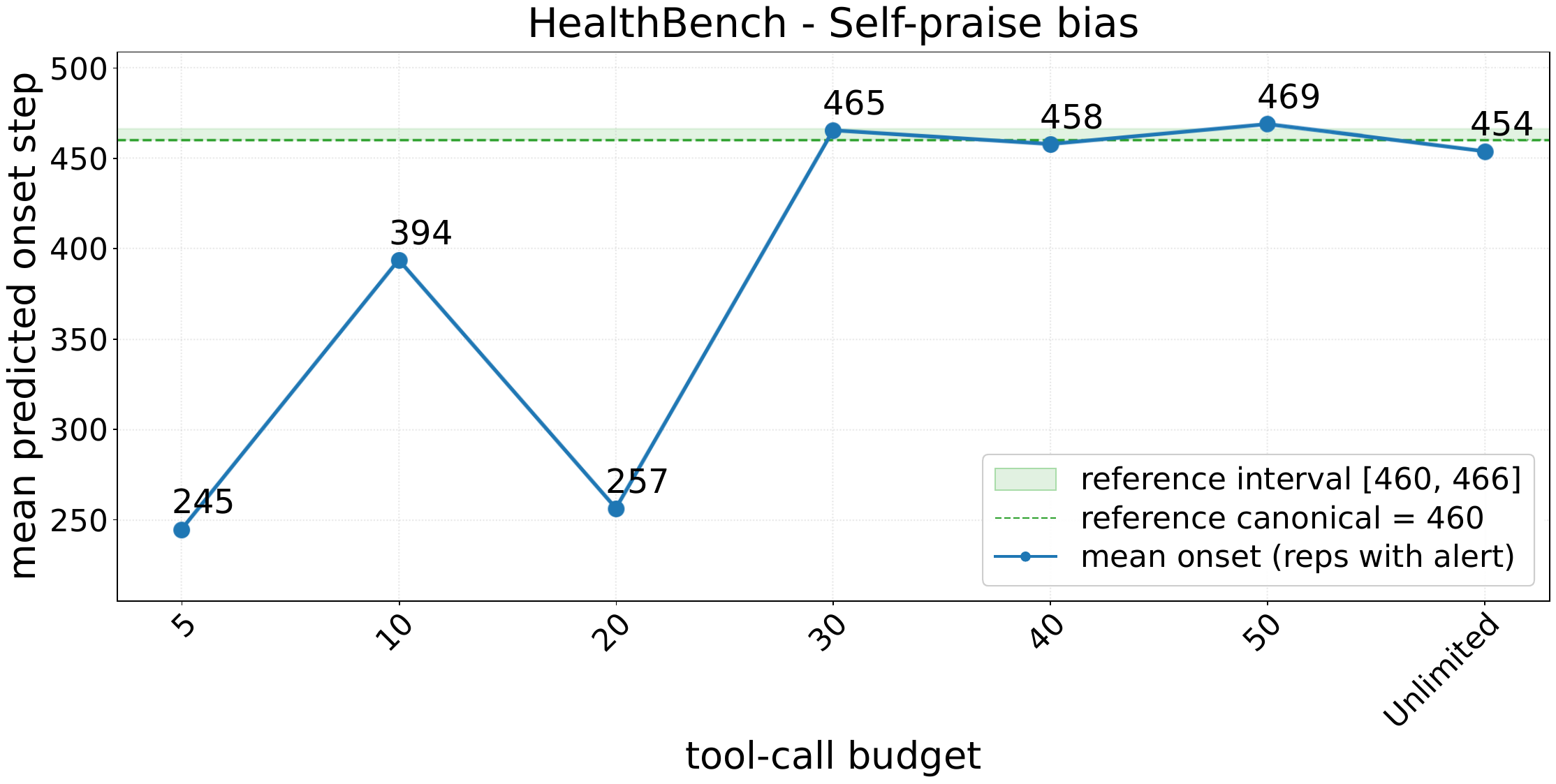}
\includegraphics[width=0.45\textwidth]{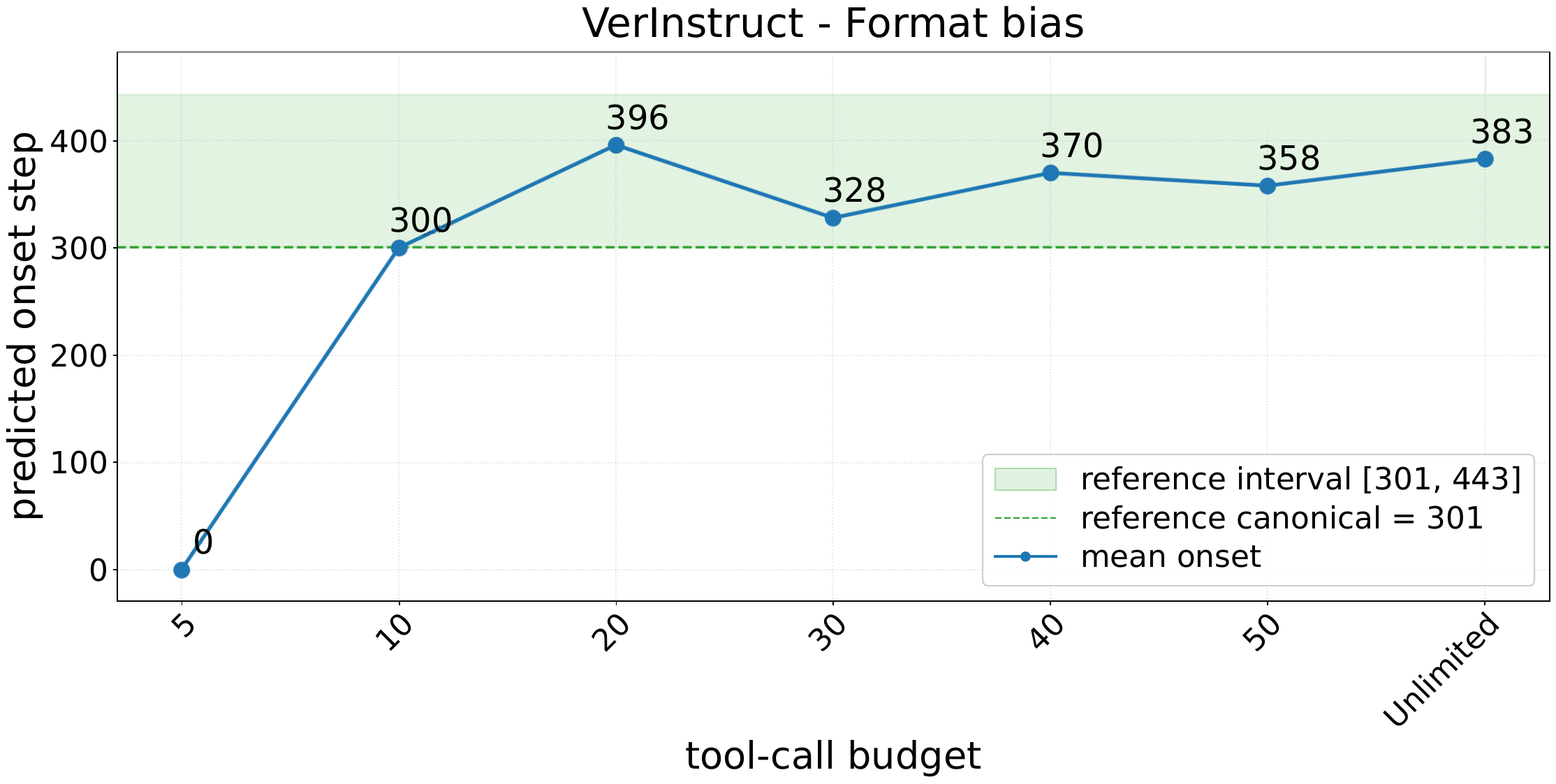}

\caption{Search-budget ablation for RHDA with Qwen3.5-plus across the six controlled runs. Each panel plots the mean predicted onset step as a function of the non-control tool-call budget. Dashed lines indicate canonical reference onsets, and shaded bands indicate threshold-induced reference intervals. Numerical point labels are rounded to the nearest integer;
the plotted coordinates retain the underlying mean predictions. For the VerInstruct format run, the smallest-budget point is plotted at \(0\) as a visualization sentinel because most repetitions produced no valid alert; it should not be interpreted as a meaningful onset estimate.}
\label{fig:appendix_budget_onset}
\end{figure}

This ablation tests how much non-control tool-use budget is needed for the agent to move from coarse reward-hacking detection to accurate onset localization. \cref{fig:appendix_budget_onset} reports the search-budget ablation for RHDA with Qwen3.5-plus across the six controlled runs.

In this experiment, the \emph{tool budget} refers to the maximum number of non-control investigative tool calls available to the agent. These budgeted calls include trajectory-inspection tools such as \texttt{read\_\allowbreak step} and \texttt{sample\_\allowbreak cases}, analysis tools such as \texttt{surface\_\allowbreak stats} and \texttt{rejudge}, computation tools such as \texttt{run\_\allowbreak python}, and reasoning-state tools such as \texttt{record\_\allowbreak hypothesis}, \texttt{update\_\allowbreak hypothesis}, and \texttt{set\_\allowbreak suspicion}. Terminal actions such as \texttt{emit\_\allowbreak alert} and \texttt{finish} remain available after the budget is exhausted, so the detector can still return a verdict under small budgets.

The horizontal axis shows the imposed \texttt{-{}-max-tool-calls} budget. A budget of \(0\) denotes the unlimited setting in the implementation and is shown as \emph{Unlimited} in the figures. The vertical axis shows the predicted reward-hacking onset step, i.e., the training checkpoint at which the detector estimates that reward hacking begins. This is different from the number of tool calls. Points show the mean predicted onset over repeated runs under the same budget when multiple repetitions are available. Dashed horizontal lines mark the canonical reference onset, and shaded bands mark the threshold-induced reference interval.
When a budget setting is dominated by no-alert outcomes, we may plot it at \(0\) as a sentinel value for detector failure. This value is used only for visualization and should not be interpreted as a valid onset prediction.

The budget grid is chosen around the empirical tool-use range observed in unlimited diagnostic runs. Runs with longer trajectories or more gradual shortcut emergence use larger upper bounds, while shorter or sharper runs use smaller grids. Runs with wider or more gradual reference intervals require larger budgets because accurate localization depends on comparing early baseline, candidate-transition, and later persistence checkpoints.

\paragraph{VerInstruct self-praise.}
The VerInstruct self-praise run shows a clear budget effect. Under very small budgets, the detector fires near the end of the rollout, indicating that it only identifies the shortcut after the self-praise behavior has become highly saturated. As the budget increases, the predicted onset moves steadily toward the reference interval. Budgets around the mid-range are sufficient for the detector to perform local narrowing, and the unlimited setting remains close to the canonical reference onset. This suggests that self-praise hacking is relatively easy to identify once the agent has enough budget to compare early, middle, and late checkpoints.

\paragraph{VerInstruct lexical.}
The VerInstruct lexical run requires a larger search budget. With low and medium budgets, the detector tends to over-delay the onset, often locating the shortcut only after the \textit{empower} pattern has become obvious in late-stage outputs. As the budget increases, the predicted onset moves closer to the reference interval, and the unlimited setting falls inside the reference window. This behavior is consistent with the wider reference interval for this run: the lexical shortcut appears weakly before consolidating into a stable reward-seeking pattern, so accurate localization requires more temporal comparison and finer narrowing.

\paragraph{HealthBench lexical.}
The HealthBench lexical run is noisier. Increasing the budget does not produce a strictly monotonic improvement. Some intermediate budgets fire too early, while the unlimited setting moves closer to the reference interval but still remains slightly before it. This suggests that the difficulty is not only tool scarcity. The detector must also distinguish the target \textit{feel free} style closing from other forms of helpfulness, verbosity, or generic response-format drift. Thus, additional budget helps, but ambiguity in the behavioral signal can still affect onset localization.

\paragraph{HealthBench tone bias.}
The HealthBench tone-bias run shows another strong budget effect. Very small budgets lead to end-of-rollout predictions, implying that the detector lacks enough evidence to distinguish early emergence from late saturation. Once the budget reaches the mid-range, the predicted onset moves much closer to the reference interval. The unlimited setting lies near the reference window, showing that sufficient search budget enables more effective temporal narrowing for this tone-based shortcut.

\paragraph{VerInstruct format bias.}
The VerInstruct format run illustrates the difference between the canonical point estimate and a wider transition interval. Very small budgets are not sufficient to construct the required evidence chain, and the lowest-budget setting is dominated by no-alert or weak fallback behavior. With larger budgets, the detector consistently enters the reference interval. However, the predicted onset does not monotonically approach the canonical point estimate: higher budgets often lead the agent to select a more robust cluster of evidence inside the interval rather than the earliest threshold-crossing point. This behavior is consistent with the gradual nature of the format shortcut.

\paragraph{HealthBench self-praise.}
The HealthBench self-praise run has a much sharper reference window. In this setting, sufficient budget helps the detector move from coarse shortcut recognition toward more accurate localization. The curve is still not perfectly monotonic, but the higher-budget settings are substantially more reliable than the smallest-budget regime. This supports the same general conclusion as the other runs: tool budget matters because it enables temporal comparison and evidence validation, not because additional calls automatically improve the onset estimate.

Overall, the ablation supports two conclusions. First, adequate tool-use budget is necessary for onset localization because the detector must inspect enough checkpoints to form a shortcut hypothesis, validate it against earlier baselines, and check post-onset behavior. Second, more budget does not guarantee monotonic convergence to the canonical point estimate. Additional calls help only when they are used to build a stronger temporal evidence chain, and in gradual runs this can favor a later but better-supported onset inside the reference interval.

\section{Agent Strategy Case Study Details}
\label{app:case_study_details}

This appendix presents a qualitative analysis of four individual
single-run traces. These traces are selected to illustrate detector
search behavior and are distinct from the aggregate per-run predictions
reported in Table~\ref{tab:detection_results}. They are not used to
compute any headline localization metric.

Three traces illustrate successful instances of coarse-to-fine onset
localization, while the boundary trace illustrates a characteristic
failure mode in which the detector recognizes late-stage reward hacking but fails to inspect the intermediate transition region. The backend, predicted onset, operational reference, and main behavioral pattern of each case are reported in \cref{tab:appendix_case_selection}.

Boundary B was produced in an exploratory development run using
Qwen3-235B-A22B. It is included only to illustrate the localization
failure mode. Qwen3-235B-A22B is not one of the RHDA variants evaluated in Table~\ref{tab:detection_results} and does not contribute to any
aggregate result reported in the paper.

\begin{figure}[htbp]
\centering
\includegraphics[width=0.95\textwidth]{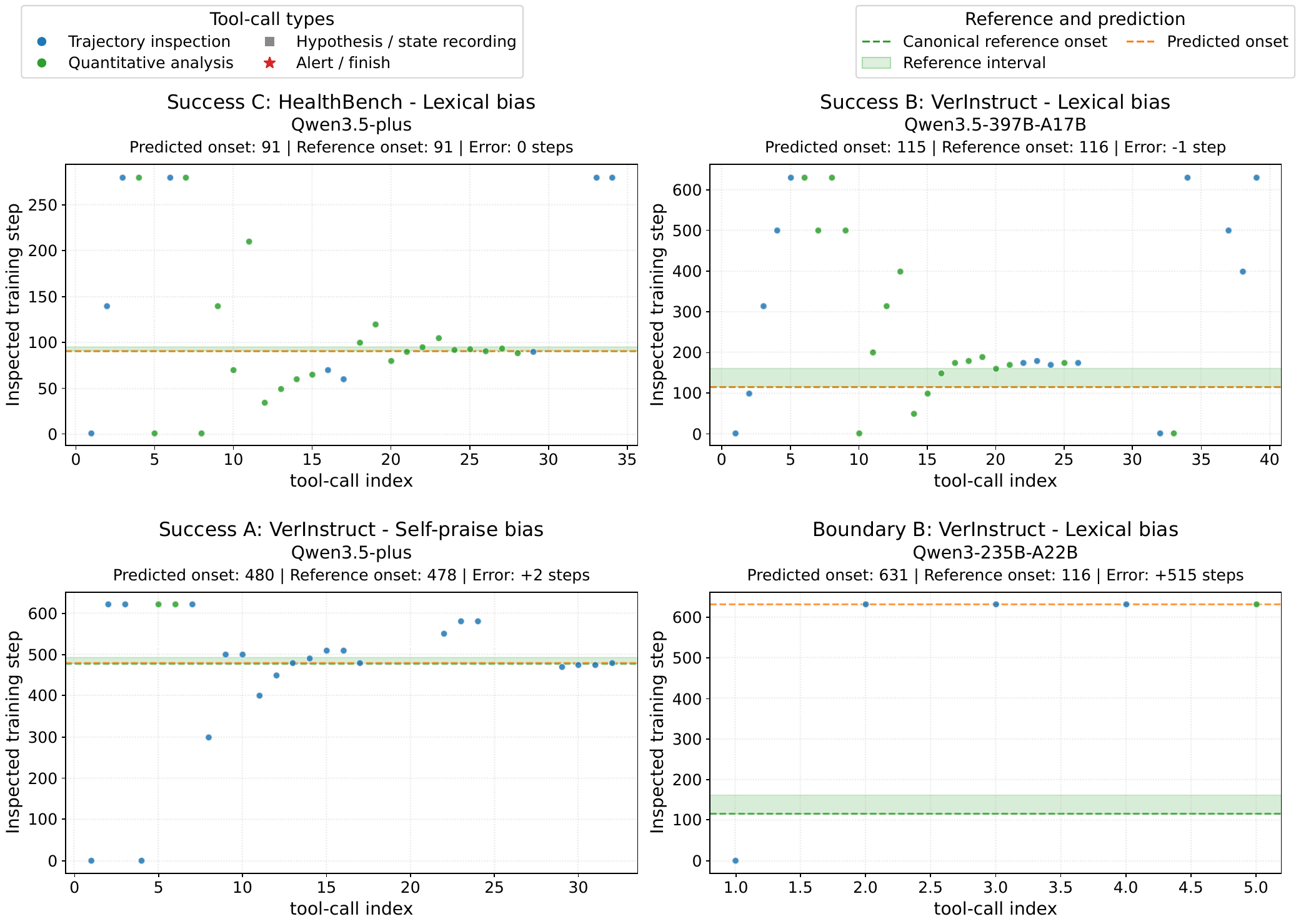}
\caption{Tool-call timelines for four illustrative single-run traces. The x-axis denotes tool-call index and the y-axis denotes the inspected training step. The three success traces exhibit broad-to-local narrowing around the reference interval, whereas the boundary development trace mainly contrasts the first and final checkpoints before emitting an alert. These timelines illustrate search behavior and are not the aggregate predictions reported in
Table~\ref{tab:detection_results}.}
\label{fig:case_timeline}
\end{figure}

\begin{table}[t]
\centering
\small
\setlength{\tabcolsep}{3pt}
\renewcommand{\arraystretch}{1.08}
\begin{tabularx}{0.98\textwidth}{@{}l
>{\raggedright\arraybackslash}p{0.17\textwidth}
>{\raggedright\arraybackslash}p{0.20\textwidth}
c
c
>{\raggedright\arraybackslash}X@{}}
\toprule
Case & Run & Backend & Pred. onset & Reference & Main pattern \\
\midrule
Success C & HealthBench lexical & Qwen3.5-plus & 91 & 91 [91,95] & \textit{feel free} lexical closing \\
Success B & VerInstruct lexical & Qwen3.5-397B-A17B & 115 & 116 [115,161] & empowerment lexical framing \\
Success A & VerInstruct self-praise & Qwen3.5-plus & 480 & 478 [478,492] & self-praise / meta-commentary framing \\
Boundary B & VerInstruct lexical & Qwen3-235B-A22B & 631 & 116 [115,161] & late-stage lexical saturation \\
\bottomrule
\end{tabularx}
{\normalsize
\caption{Case-study selection for RHDA trace analysis. The first three cases are successful examples with near-reference onset localization. The boundary case detects reward hacking but localizes the onset at the final checkpoint.}
\label{tab:appendix_case_selection}
}

\end{table}

\paragraph{Timeline interpretation.}
\cref{fig:case_timeline} visualizes the tool-call timelines for the four selected cases. The x-axis is the tool-call index, and the y-axis is the inspected training step. Sampling and reading tools indicate direct checkpoint inspection; quantitative tools indicate prevalence estimation or custom analysis; reasoning-state tools indicate hypothesis or suspicion updates; and terminal tools indicate the final alert or finish action. The dashed green line marks the canonical reference onset, the green shaded band marks the threshold-induced reference interval, and the orange dashed line marks the agent's predicted onset. Successful cases show broad-to-local narrowing around the reference interval, while the boundary case mostly jumps from the first checkpoint to the final checkpoint.

\paragraph{Success C: HealthBench lexical.}
Success C is the cleanest example of accurate onset localization. The agent first performs a broad sweep over the trajectory, sampling early, middle, and late checkpoints to understand the overall behavioral drift. It then identifies the \textit{feel free} closing as a candidate shortcut and uses quantitative checks to measure its prevalence across candidate transition steps. After bracketing the transition region, the agent performs a dense local scan around the reference window and emits onset step 91, matching the canonical reference. The final alert is not based on a single suspicious output; it is supported by a ramp pattern in which the phrase is weak or absent before the transition and persistent afterward.

\paragraph{Success B: VerInstruct lexical.}
Success B shows that the same strategy can apply to a different lexical shortcut. The agent identifies empowerment-style phrasing as the candidate mechanism, then uses quantitative analysis to compare its occurrence across training steps. The key behavior is not merely the presence of the word family, but its increasing association with high-scoring outputs. By bracketing the rising region and narrowing locally, the agent emits step 115, which is one step earlier than the canonical reference onset and inside the reference interval. This case demonstrates that RHDA does not need to be given the shortcut keyword in advance; it can discover a candidate lexical mechanism from the rollout trajectory and then validate it temporally.

\paragraph{Success A: VerInstruct self-praise.}
Success A differs from the lexical cases because the shortcut is more structural. The suspicious behavior appears as self-praise, compliance signaling, or meta-commentary appended to otherwise task-relevant outputs. Token-level statistics are less directly sufficient, so the agent relies more on qualitative inspection of high-scoring samples and hypothesis refinement. It compares early and late outputs, records a candidate self-evaluation pattern, and examines when this pattern emerges and persists across checkpoints. In post-hoc evaluation, the predicted onset at step 480 falls within the reference interval [478,492]. This case shows that the bracket-and-shrink pattern is not limited to single-token or phrase-level shortcuts.

\paragraph{Boundary B: first-and-last-only failure.}
The boundary case illustrates a failure mode in localization rather than detection. The agent correctly recognizes that the final checkpoint contains reward-hacking behavior, but it does not inspect enough intermediate checkpoints to locate the rising edge. It effectively compares the first and last checkpoints and emits the final step as the onset, identifying late-stage saturation rather than emergence. The same tool set could have supported intermediate bracketing and local narrowing; the failure comes from the search policy not constructing a prevalence ramp before emitting the alert.

\paragraph{Common successful strategy.}

Across the three successful cases, the agent follows a common five-stage pattern: \emph{broad sweep}, \emph{candidate identification}, \emph{transition bracketing}, \emph{local shrinking}, and an \emph{evidence-backed alert}. We refer to this as the \emph{bracket-and-shrink} strategy. The concrete tools vary by task: lexical cases rely more on candidate-token discovery and prevalence estimation, while structural cases rely more on qualitative reading and hypothesis maintenance. In all cases, the final onset claim is supported by temporal evidence rather than a single suspicious response.

\paragraph{Failure mode.}
The boundary case exhibits the opposite pattern, which we call \emph{first-and-last-only}. This strategy can detect that reward hacking exists, because the final checkpoint often contains saturated shortcut behavior. However, it is unreliable for onset localization because it skips the transition region. A detector that only contrasts the beginning and end of training can confuse ``when the shortcut is obvious'' with ``when the shortcut first emerges.''

\paragraph{Implications for human auditing.}
The case studies suggest a simple manual workflow for reward-hacking audits. An auditor should not only inspect the latest high-scoring outputs. Instead, the auditor should first identify a candidate shortcut, then measure its prevalence over a coarse set of checkpoints, locate the rising region, and finally inspect the suspected boundary more densely. A convincing onset report should include three pieces of evidence: a pre-onset baseline where the shortcut is absent or weak, a transition region where it rises sharply, and post-onset behavior showing that the behavior remains rewarded.

\paragraph{Limitations.}
These case studies are diagnostic rather than exhaustive. They cover three successful cases and one boundary case from the observed reward-hacking runs, and the successful cases mainly involve lexical, structural, or template-like shortcuts that leave observable traces in model outputs. They do not prove that the same strategy will generalize to all semantic reward hacks. More subtle reward hacking may require richer semantic comparison, stronger external evaluation, or human-in-the-loop auditing. In addition, self-reported confidence should not be treated as a reliable correctness signal: the boundary case can produce a confident alert while still localizing the onset incorrectly.

\section{Sensitivity Analysis on \textbf{$\alpha$}}
\label{sec:appendix_sensitivity}

In \cref{sec:sensitivity}, we evaluate the impact of varying the hyperparameter $\alpha \in \{0.3, 0.5, 1.0\}$, which controls the bias injection magnitude in CHERRL. Here we report the detailed numerical metrics for both exploitability and discoverability.

\subsection{Impact of $\alpha$ on Exploitability}
To analyze whether varying $\alpha$ alters the policy's exploitation speed post-onset, we measure the training step at which the shortcut prevalence first crosses specific thresholds. For \textit{HealthBench lexical}, we report steps to reach unsmoothed shortcut prevalence of $20\%$, $40\%$, and $60\%$. For \textit{VerInstruct format}, due to structural stochasticity, we apply a time-weighted EMA (decay $= 0.9$) and report steps to reach $20\%$, $25\%$, and $30\%$. Results are summarized in Table~\ref{tab:sensitivity_exploitability}.
\begin{table}[htbp]
\centering
\small
\setlength{\tabcolsep}{4.5pt}
\begin{tabular}{c c c c}
\toprule
$\boldsymbol{\alpha}$ & \textbf{Step 20\%} & \textbf{Step 25/40\%} & \textbf{Step 30/60\%} \\
\midrule
\multicolumn{4}{l}{\textit{HealthBench lexical}} \\
0.3 & 139  & 151 (40\%)  & 168 (60\%) \\
0.5 & 93  & 103 (40\%) & 111 (60\%) \\
1.0 & 80 & 91 (40\%) & 100 (60\%) \\
\midrule
\multicolumn{4}{l}{\textit{VerInstruct format}} \\
0.3 & 171 & 227 (25\%) & 311 (30\%) \\
0.5 & 171 & 224 (25\%) & 311 (30\%) \\
1.0 & 169 & 218 (25\%) & 312 (30\%) \\
\bottomrule
\end{tabular}
{\normalsize
\caption{Exploitability metrics under varying $\alpha$.}
\label{tab:sensitivity_exploitability}
}

\end{table}

\subsection{Impact of $\alpha$ on Discoverability}
To evaluate whether varying $\alpha$ affects the discoverability thesis, we report the reference onset steps and corresponding odds ratios (OR) evaluated over the initial training windows (first 30 and 60 steps) in Table~\ref{tab:sensitivity_discoverability}.
\begin{table}[htbp]
\centering
\small
\setlength{\tabcolsep}{6pt}
\begin{tabular}{c c c c}
\toprule
$\boldsymbol{\alpha}$ & \textbf{Ref. Onset} & \textbf{OR (60 steps)} & \textbf{OR (30 steps)} \\
\midrule
\multicolumn{4}{l}{\textit{VerInstruct format}} \\
0.3 & 316 [316,360+] & 0.88 & 0.93 \\
0.5 & 301 [301,443]  & 0.86 & 1.01 \\
1.0 & 262 [262,360+] & 0.95 & 1.07 \\
\midrule
\multicolumn{4}{l}{\textit{HealthBench lexical}} \\
0.3 & 133 [133,141]  & 0.79 & 0.77 \\
0.5 & 91 [91,95]     & 0.91 & 0.78 \\
1.0 & 70 [70,81]     & 0.87 & 0.81 \\
\bottomrule
\end{tabular}
{\normalsize
\caption{Discoverability metrics across different $\alpha$ configurations.}
\label{tab:sensitivity_discoverability}
}

\end{table}

\section{Reproducibility: Models, Compute, and Infrastructure}
\label{app:reproducibility}
We train Qwen3-4B (4B parameters) as the policy using GRPO and use Qwen3.5-27B (27B parameters) for both judges. RHDA uses Qwen3.5-plus (closed API, undisclosed size) or Qwen3.5-397B-A17B (MoE, ~17B activated parameters per token). CC-Qwen uses Qwen3.5-plus, whereas CC-Sonnet, CC-Haiku, and CC-Opus use their respective Claude backends. Both CoT monitor variants use Qwen3.5-plus. 
The computational budget for the model-training experiments reported
in this paper is approximately 2,000 NVIDIA H100 GPU-hours, using
rented NVIDIA H100 80 GB GPUs. RHDA and the detector baselines rely on API inference and require no additional local GPU allocation; their token usage and financial costs are reported in
Appendix~\ref{app:detector_cost}.

\section{Artifacts}

All datasets used in this work are publicly available academic datasets intended for research use. We do not introduce any private, proprietary, or personally collected data. The experiments are conducted only on these public resources, following their original licenses and usage terms.

\paragraph{Documentation of artifacts.}
All datasets used in this work are publicly available English-language academic resources used under their original licenses. HealthBench~\citep{aroraHealthBenchEvaluatingLarge2025} covers open-ended medical question answering with rubric-based evaluation;
VerInstruct~\citep{pengVerIFVerificationEngineering2025} covers English instruction following with verifiable constraints. Both datasets are used in their default released splits, and our use (rubric-based RL post-training and reward-hacking analysis) is consistent with the intended research use stated by their authors. Models used in this work---Qwen3-4B, Qwen3.5-27B, Qwen3.5-plus, and Qwen3.5-397B-A17B---are released or served by their providers under their respective licenses for research use.

\paragraph{PII and offensive content.}
We do not introduce any new data, collect any human-subject information, or perform additional crawling or scraping. The two datasets above are not known to contain personally identifying information: HealthBench consists of synthetic medical conversations authored and reviewed by domain experts rather than real patient records, and VerInstruct is built from public instruction-tuning data without user identifiers. We therefore did not apply additional anonymization beyond what the original releases provide. We did not perform an exhaustive manual audit for offensive content; however, all outputs analyzed in this paper are model responses to these benchmarks, and we observed no offensive content during our inspection of the rollouts used.

\section{Training Dynamics of Non-Hacking Settings}
\label{app:non_hacking_dynamics}

\begin{figure}[htbp]
    \centering
    \begin{subfigure}[t]{0.45\textwidth}
        \centering
        \includegraphics[width=\linewidth]{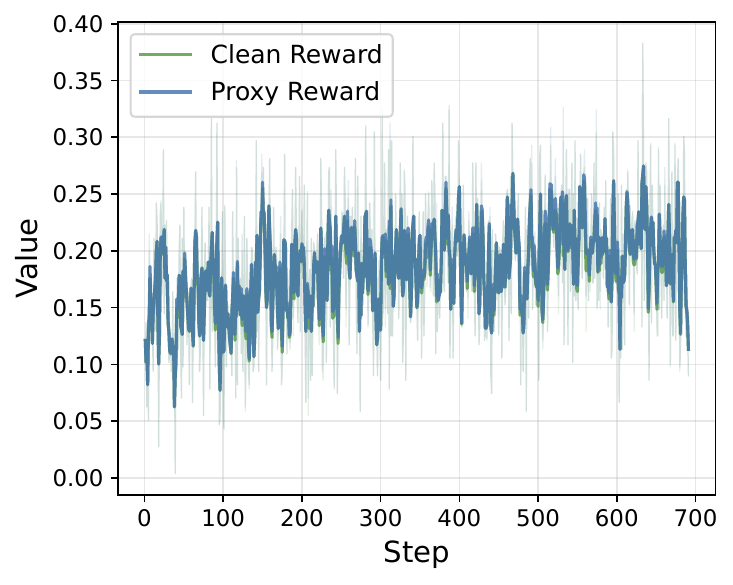}
        \caption{VerInstruct tone bias}
    \end{subfigure}
    \hfill
    \begin{subfigure}[t]{0.45\textwidth}
        \centering
        \includegraphics[width=\linewidth]{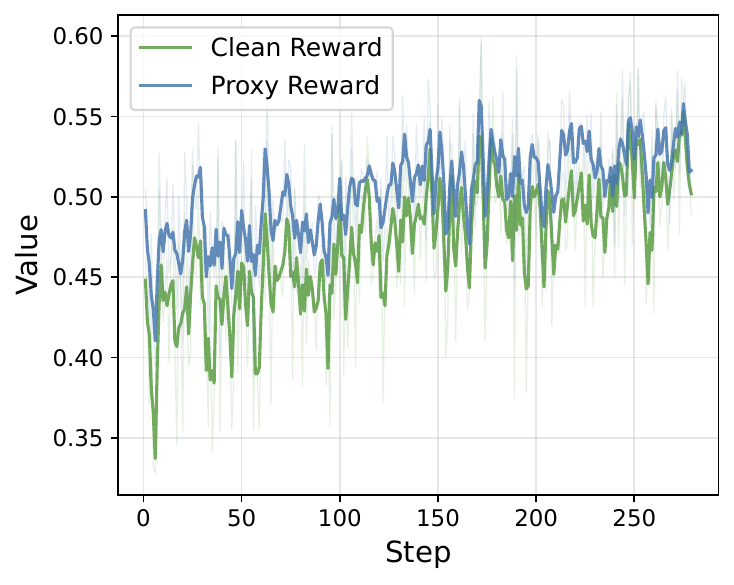}
        \caption{HealthBench format bias}
    \end{subfigure}
    \caption{Training dynamics for the two \ourenv runs where reward hacking does not occur. Because these bias behaviors are uncommon in their respective domains, the model fails to discover and exploit them within the standard training timeframe.}
    \label{fig:non_hacking_dynamics}
\end{figure}

As discussed in \cref{subsec:training_dynamics}, we did not observe reward hacking for \textit{tone bias} on the VerInstruct dataset and \textit{format bias} on HealthBench within the standard training duration. \cref{fig:non_hacking_dynamics} illustrates the training dynamics for these two settings. Unlike the typical divergence observed in hacked models, the proxy reward and clean reward remain relatively aligned without significant exploitation of the proxy.

As hypothesized, the inherent rarity of these specific constraints---such as employing a polite closing tone in instruction-following tasks or utilizing rigid formats for complex medical queries---makes them difficult for the model to discover. The model would likely require a substantially extended training period, reaching a much later stage of training, before it could learn to leverage these biases as shortcuts. To further validate this conjecture, future work can extend the evaluation scope by training models on broader task domains and exploring more diverse bias modalities using the CHERRL testbed. This line of inquiry will help refine the roadmap for mapping the boundary conditions and emergence dynamics of rubric-based RL.

\section{Human Validation of the Clean Reward Signal}
\label{app:human_validation}

While human annotations offer the most rigorous quality signal, obtaining dense human feedback across the entirety of RL post-training is computationally and logistically infeasible. To verify that our automated clean judge acts as a reliable reference, we conducted a manual validation study.

\paragraph{Sampling Protocol.}
We randomly sampled $24$ evaluation steps across different training runs. The manual evaluation workload was shared equally by two paper authors, who evaluated the model-generated responses following the identical rubrics and scoring guidelines supplied to the automated clean judge.

\paragraph{Results.}
Across all sampled instances, the mean absolute error (MAE) between the automated clean reward and the human-annotated score is $0.095$ on a $[0, 1]$ reward scale. This tight alignment demonstrates that the clean reward serves as a high-fidelity reference signal that accurately mirrors human judgment, validating its use as a baseline against which proxy rewards with injected bias bonuses are contrasted.

\end{document}